\documentclass{vldb-arxiv}
\usepackage{graphicx}
\usepackage{balance}  

\usepackage{booktabs} 
\usepackage[rounded]{syntax}
\usepackage{amsmath}
\usepackage{tikz}
\usetikzlibrary{positioning,automata}
\usepackage{xhfill}
\usepackage{upquote}
\usepackage{graphicx}
\usepackage{textcomp}
\usepackage{enumerate}
\usepackage{caption}

\usepackage{forest}
\usepackage{multirow}
\usepackage{fancybox}
\usepackage{flushend}
\usepackage{graphicx} 
\newcommand{\squishlist}{
 \begin{list}{$\bullet$}
  { \setlength{\itemsep}{0pt}
     \setlength{\parsep}{3pt}
     \setlength{\topsep}{3pt}
     \setlength{\partopsep}{0pt}
     \setlength{\leftmargin}{1.0em}
     \setlength{\labelwidth}{1em}
     \setlength{\labelsep}{0.5em} } }

\newcommand{\squishend}{
  \end{list}  }

\vldbTitle{Semantically Driven Auto-completion}
\vldbAuthors{Konstantine Arkoudas, and Mohamed Yahya}
\vldbDOI{https://doi.org/10.14778/xxxxxxx.xxxxxxx}
\vldbVolume{12}
\vldbNumber{xxx}
\vldbYear{2019}

\begin{document}


\title{Semantically Driven Auto-completion}



%
%
%
%

\numberofauthors{2} 

\author{
%
%
\alignauthor
Konstantine Arkoudas\\
       \affaddr{Bloomberg}\\
       \affaddr{New York}\\
       \affaddr{USA}\\
       \email{karkoudas@bloomberg.net}
\alignauthor
Mohamed Yahya\\
       \affaddr{Bloomberg}\\
       \affaddr{London}\\
       \affaddr{United Kingdom}\\
       \email{myahya6@bloomberg.net}
}

\maketitle

\begin{abstract}
The Bloomberg Terminal has been a leading source of
financial data and analytics for over 30 years.
Through its thousands of functions, the Terminal allows its users 
to query and run analytics over a large array of 
data sources, including structured, semi-structured, and unstructured
data; as well as plot charts, set up event-driven alerts and triggers,
create interactive maps, exchange information via instant and email-style
messages, and so on. 
To improve user experience,
we have been building question answering systems that can understand
a wide range of natural language constructions for various domains
that are of fundamental interest to our users. 
Such natural language interfaces, while
exceedingly helpful to users,  introduce a number of usability
challenges of their own. We tackle some of these challenges
through auto-completion for query formulation.
A distinguishing mark of our auto-complete systems is that they
are based on and guided by corresponding
semantic parsing systems. We describe the auto-complete problem
as it arises in this setting, the novel algorithms that
we use to solve it, and report on the quality of the results 
and the efficiency of our approach. 
\end{abstract}


\newcommand{\restrict}{\mbox{$\restriction$}}
\newcommand{\restrictsp}{\mbox{$\restriction$} }
\newcommand{\ar}{\mbox{\em r}}
\newcommand{\arsp}{\mbox{\em r} }
\newcommand{\arity}{\mbox{\em r}}
\newcommand{\aritysp}{\mbox{\em r} }
\newcommand{\dom}[1]{\mbox{$\temls{Dom}(#1)$}}
\newcommand{\domsp}{\mbox{\em Dom\/}}
\newcommand{\ran}{\mbox{\em Ran}}
\newcommand{\ransp}{\mbox{\em Ran }}
\newcommand{\Bool}{\mbox{\bf Bool}}
\newcommand{\Boolsp}{\mbox{\bf Bool} }
\newcommand{\eps}{\mbox{$\epsilon$}}
\newcommand{\epsp}{\mbox{$\epsilon$} }
\newcommand{\sig}{\mbox{$\sigma$}}
\newcommand{\sigsp}{\mbox{$\sigma$} }
\newcommand{\Sig}{\mbox{$\Sigma$}}
\newcommand{\Sigsp}{\mbox{$\Sigma$} }
\newcommand{\Gam}{\mbox{$\Gamma$}}
\newcommand{\Gamsp}{\mbox{$\Gamma$} }
\newcommand{\gam}{\mbox{$\gamma$}}
\newcommand{\gamsp}{\mbox{$\gamma$} }
\newcommand{\mrho}{\mbox{$\rho$}}
\newcommand{\mrhosp}{\mbox{$\rho$ }}
\newcommand{\msigma}{\mbox{$\sigma$}}
\newcommand{\msigmasp}{\mbox{$\sigma$} }
\newcommand{\mtheta}{\mbox{$\theta$}}
\newcommand{\mthetasp}{\mbox{$\theta$ }}
\newcommand{\al}{\mbox{$\alpha$}}
\newcommand{\bet}{\mbox{$\beta$}}
\newcommand{\bett}{\mbox{$\beta$ }}
\newcommand{\mlam}{\mbox{$\lambda$}}
\newcommand{\mlamsp}{\mbox{$\lambda$} }
\newcommand{\Lam}{\mbox{$\Lambda$}}
\newcommand{\Lamt}{\mbox{$\Lambda$ }}
\newcommand{\negsp}{\mbox{$\neg$ }}
\newcommand{\negnsp}{\mbox{$\neg$}}
\newcommand{\mand}{\mbox{$\:\wedge \:$}}
\newcommand{\mandsp}{\mbox{$\wedge$ }}
\newcommand{\mor}{\mbox{$\:\vee \:$}}
\newcommand{\morsp}{\mbox{$\vee$ }}
\newcommand{\mif}{\mbox{$\:\Rightarrow\,$}}
\newcommand{\mifsp}{\mbox{$\:\Rightarrow\,$ }}
\newcommand{\miff}{\mbox{$\:\Leftrightarrow \,$}}
\newcommand{\miffsp}{\mbox{$\Leftrightarrow$ }}
\newcommand{\mnegnsp}{\mbox{$\neg$}}
\newcommand{\mandnsp}{\mbox{$\wedge$}}
\newcommand{\mornsp}{\mbox{$\vee$}}
\newcommand{\mifnsp}{\mbox{$\Rightarrow$}}
\newcommand{\miffnsp}{\mbox{$\Leftrightarrow$}}
\newcommand{\at}{\mbox{\em ATOM}}
\newcommand{\tvs}{\mbox{$\{${\bf T, F}$\}\:$}}
\newcommand{\tv}{\mbox{${\mathcal TV}$}}
\newcommand{\True}{\mbox{\bf T}}
\newcommand{\False}{\mbox{\bf F}}
\newcommand{\true}{\mbox{\bf true}}
\newcommand{\false}{\mbox{\bf false}}
\newcommand{\strue}{\mbox{\bf t}}
\newcommand{\sfalse}{\mbox{\bf f}}
\newcommand{\struesp}{\mbox{\bf t} }
\newcommand{\sfalsesp}{\mbox{\bf f} }
\newcommand{\Truesp}{\mbox{\bf T }}
\newcommand{\Falsesp}{\mbox{\bf F }}
\newcommand{\truesp}{\mbox{\bf true }}
\newcommand{\falsesp}{\mbox{\bf false }}
\newcommand{\sposet}{\mbox{$(A,\prec)$ }}
\newcommand{\wposet}{\mbox{$(A,\preceq)$ }}
\newcommand{\ssqposet}{\mbox{$(A,\sqpsub)$ }}
\newcommand{\wsqposet}{\mbox{$(A,\sqsub)$ }}
\newcommand{\rarrs}{\mbox{$\;\rightarrow\;$}}
\newcommand{\rarrls}{\mbox{$\,\rightarrow\:$}}
\newcommand{\dhrarr}{\mbox{$\twoheadrightarrow$}}
\newcommand{\dhrarrs}{\mbox{$\;\twoheadrightarrow\;$}}
\newcommand{\dhrarrls}{\mbox{$\,\twoheadrightarrow\,$}}
\newcommand{\larrls}{\mbox{$\,\leftarrow\;$}}
\newcommand{\rarrstarls}{\mbox{$\,\rightarrow^*\,$}}
\newcommand{\rarrcrossls}{\mbox{$\,\rightarrow+*\,$}}
\newcommand{\rarr}{\mbox{$\rightarrow$}}
\newcommand{\rarrt}{\mbox{$\rightarrow$} }
\newcommand{\Rarr}{\mbox{$\Rightarrow$}}
\newcommand{\Rarrls}{\mbox{$\,\Rightarrow\,$}}
\newcommand{\Rarrtc}{\mbox{$\,\Rightarrow^+ \,$}}
\newcommand{\Rarrtrc}{\mbox{$\,\Rightarrow^* \,$}}
\newcommand{\larr}{\mbox{$\leftarrow$}}
\newcommand{\Larr}{\mbox{$\Leftarrow$}}
\newcommand{\hrarr}{\mbox{$\hookrightarrow$}}
\newcommand{\hrarrsp}{\mbox{$\hookrightarrow$ }}
\newcommand{\hrarrls}{\mbox{$\:\hookrightarrow\:$}}
\newcommand{\sub}{\mbox{$\; \subseteq \;$}}
\newcommand{\psub}{\mbox{$\; \subset \;$}}
\newcommand{\super}{\mbox{$\; \supseteq \;$}}
\newcommand{\psuper}{\mbox{$\; \supset \;$}}
\newcommand{\lrarr}{\mbox{$\leftrightarrow$}}
\newcommand{\lrarrls}{\mbox{$\:\leftrightarrow\:$}}
\newcommand{\Lrarr}{\mbox{$\Leftrightarrow$}}
\newcommand{\lRarr}{\mbox{$\Longrightarrow$}}
\newcommand{\lRarrsp}{\mbox{$\Longrightarrow$ }}
\newcommand{\lRarrls}{\mbox{$\,\Longrightarrow\,$}}
\newcommand{\sep}{\mbox{$\; | \;$}}
\newcommand{\sepls}{\mbox{$\:\, | \:$}}
\newcommand{\sepms}{\mbox{$\; | \;$}}
\newcommand{\sats}{\mbox{$\, \models \,$}}
\newcommand{\nsats}{\mbox{$\, \not\models \,$}}
\newcommand{\nullset}{\mbox{$\emptyset\,\:$}}
\newcommand{\eset}{\mbox{$\emptyset$}}
\newcommand{\esetsp}{\mbox{$\emptyset$ }}
\newcommand{\sqsub}{\mbox{$\; \sqsubseteq \;$}}
\newcommand{\sqpsub}{\mbox{$\; \sqsubset \;$}}
\newcommand{\sqsup}{\mbox{$\; \sqsupseteq \;$}}
\newcommand{\sqpsup}{\mbox{$\; \sqsupset \;$}}
\newcommand{\bsquare}{\rule{2.2mm}{2.2mm}}
\newcommand{\bsig}{\mbox{$\;\bsquare$}}
\newcommand{\inc}{\mbox{\em s}}
\newcommand{\incsp}{\mbox{\em s} }
\newcommand{\tar}{\mbox{${\mathcal T}\hspace*{-0.017in}{\mathcal A}{\mathcal R}$}}
\newcommand{\tarsp}{\mbox{${\mathcal T}\hspace*{-0.017in}{\mathcal A}{\mathcal R}$} }
\newcommand{\zero}{\mbox{\em 0}}
\newcommand{\zerosp}{\mbox{\em 0 }}
\newcommand{\plus}{\mbox{\em plus}}
\newcommand{\plusp}{\mbox{\em plus }}
\newcommand{\mult}{\mbox{\em times}}
\newcommand{\multsp}{\mbox{\em times }}
\newcommand{\pset}[1]{\mbox{$\mathcal{P}(#1)$}}
\newcommand{\psetsp}[1]{\mbox{$\mathcal{P}(#1)$} }
\newcommand{\finpset}[1]{\mbox{$\mathcal{P}_{\infty}(#1)$}}
\newcommand{\finpsetsp}[1]{\mbox{$\mathcal{P}_{\infty}(#1)$} }
\newcommand{\ded}{\mbox{$\mathcal{D}$}}
\newcommand{\dedsp}{\mbox{$\mathcal{D}$} }
\newcommand{\lttp}{\mbox{\tt (}}
\newcommand{\lttpsp}{\mbox{\tt (} }
\newcommand{\rttp}{\mbox{\tt )}}
\newcommand{\rttpsp}{\mbox{\tt )} }
\newcommand{\mc}[1]{\mbox{$\mathcal{#1}$}}
\newcommand{\fmc}[1]{\mbox{$\footnotesize\mathcal{#1}$}}
\newcommand{\scmc}[1]{\mbox{$\scriptstyle\mathcal{#1}$}}
\newcommand{\beq}[2]{\begin{equation} #1 \label{#2} \end{equation}}
\newcommand{\irule}[3]{
\begin{tabular}{cl} 
$\;\;$#1$\;$ & \mbox{}  \\[-0.099in] \hrulefill & \hspace*{-0.08in} \mbox{$\;\:$#3} \\[-0.03in] 
$\;\;$#2$\;$ & \mbox{} 
\end{tabular}}

\newcommand{\lirule}[3]{
\begin{tabular}{rc} 
\mbox{} & $\;\;$#1$\;$  \\[-0.099in] 
#3 \hspace*{-0.08in} & \hrulefill \\[-0.03in] 
\mbox{} & $\;\;$#2$\;$ 
\end{tabular}}

\newcommand{\extrairule}[4]{
\begin{tabular}{cl} 
\begin{tabular}{c}
$\;\;$#1$\;$ \\[-0.099in] 
\hrulefill  \\[-0.03in] $\;\;$#2$\;$ \end{tabular} & \hspace*{-0.2in} #3 \\
#4 & \mbox{} 
\end{tabular}}

\newcommand{\extralirule}[6]{
\begin{tabular}{rc}  
#3 \hspace*{#4} & 
\begin{tabular}{c}
$\;\;$#1$\;$ \\[-0.099in] 
\hrulefill  \\[-0.03in] $\;\;$#2$\;$ \end{tabular} \\[#6]
\mbox{} & #5 
\end{tabular}}

\newcommand{\mtt}[1]{\mbox{\tt #1}}
\newcommand{\smtt}[1]{\mbox{\small\tt #1}}
\newcommand{\fmtt}[1]{\mbox{\footnotesize\tt #1}}
\newcommand{\scmtt}[1]{\mbox{\scriptsize\tt #1}}
\newcommand{\mbf}[1]{\mbox{\bf #1}}
\newcommand{\mforall}{\mbox{$\forall$}}
\newcommand{\mforallsp}{\mbox{$\forall$} }
\newcommand{\mexists}{\mbox{$\exists$}}
\newcommand{\existsUnique}{\mbox{$\exists!$}}
\newcommand{\mexistsp}{\mbox{$\exists$} }
\newcommand{\freevar}{\mbox{\em FV\/}}
\newcommand{\freevarls}{\mbox{\em FV}\hspace*{0.006in}}
\newcommand{\freevarsp}{\mbox{\em FV\/} }
\newcommand{\boundvar}{\mbox{\em BV}}
\newcommand{\boundvarls}{\mbox{\em BV}\hspace*{0.006in}}
\newcommand{\boundvarsp}{\mbox{\em BV\/} }
\newcommand{\be}{\begin{enumerate}}
\newcommand{\ee}{\end{enumerate}}
\newcommand{\bi}{\begin{itemize}}
\newcommand{\ei}{\end{itemize}}
\newcommand{\bit}{\begin{itemize}}
\newcommand{\eit}{\end{itemize}}
\newcommand{\ben}{\begin{enumerate}}
\newcommand{\een}{\end{enumerate}}
\newcommand{\tem}[1]{\mbox{\em #1}}
\newcommand{\temv}[1]{\mbox{\em #1\/}}
\newcommand{\temls}[1]{\mbox{\em #1}\hspace*{0.006in}}
\newcommand{\temts}[1]{\mbox{\em #1}\hspace*{0.009in}}
\newcommand{\temTMode}[1]{\mbox{#1}}
\newcommand{\temvTMode}[1]{\mbox{#1\/}}
\newcommand{\temlsTMode}[1]{\mbox{#1}\hspace*{0.006in}}
\newcommand{\mcup}{\mbox{$\cup$}}
\newcommand{\mcupsp}{\mbox{$\cup$ }}
\newcommand{\mcupls}{\mbox{$\:\cup\:$}}

\newcommand{\tterms}[3]{\mbox{$\mbf{Terms}(#1,#2,#3)$}}

\newcommand{\grterms}[1]{\mbox{$\mbf{Terms}(#1)$}}

\newcommand{\nats}{\mbox{$N$}}
\newcommand{\natsp}{\mbox{$N$ }}
\newcommand{\pnats}{\mbox{$N_+$}}
\newcommand{\pnatsp}{\mbox{$N_+$ }}
\newcommand{\bools}{\mbox{$B$}}
\newcommand{\boolsp}{\mbox{$B$ }}
\newcommand{\nat}{\mbox{nat}}
\newcommand{\formsub}[4]{$\mbox{#1}\hspace*{-0.04in}\overset{\small \mbox{#2}/\mbox{#3}}
{\leadsto}\hspace*{-0.04in}\mbox{#4}$}
\newcommand{\altformsub}[4]{$\mbox{#1}\leadsto\mbox{#2}\{\mbox{#3}/\mbox{#4}\}$}
\newcommand{\propsent}{\mbox{\bf PropSent}}
\newcommand{\propsentsp}{\mbox{\bf PropSent }}
\newcommand{\propat}{\mbox{\bf PropAt}}
\newcommand{\propatsp}{\mbox{\bf PropAt }}
\newcommand{\var}[1]{\mbox{\rm $\mbox{\em Var\/}(#1)$}}
\newcommand{\varls}{\mbox{\em Var}\hspace*{0.006in}}
\newcommand{\varsp}{\mbox{\em Var\/} }
\newcommand{\typsub}{\mbox{$\{v_1\mapsto t_1,\ldots,v_n \mapsto t_n\}$}}
\newcommand{\typsubsp}{\mbox{$\{v_1\mapsto t_1,\ldots,v_n \mapsto t_n\}$} }
\newcommand{\id}[1]{$\mbox{id}_{#1}$}
\newcommand{\idsp}[1]{$\mbox{id}_{#1}$ }
\newcommand{\aka}{\mbox{a.k.a. }}
\newcommand{\ie}{\mbox{i.e.} }
\newcommand{\iensp}{\mbox{i.e.}}
\newcommand{\Iensp}{\mbox{I.e.}}
\newcommand{\iesp}{\mbox{i.e. }}
\newcommand{\Ie}{\mbox{I.e.} }
\newcommand{\Iesp}{\mbox{I.e. }}
\newcommand{\etc}{\mbox{etc.}}
\newcommand{\etcsp}{\mbox{etc.} }
\newcommand{\eg}{\mbox{e.g.} }
\newcommand{\egnsp}{\mbox{e.g.}}
\newcommand{\Egnsp}{\mbox{E.g.}}
\newcommand{\Eg}{\mbox{E.g.} }
\newcommand{\egsp}{\mbox{e.g. }}
\newcommand{\Egsp}{\mbox{E.g. }}
\newcommand{\successor}{\mbox{$\,^{\prime}$}}
\newcommand{\successorsp}{\mbox{$\,^{\prime}\,$ }}
\newcommand{\mtilde}{\mbox{$\tilde{}\:$}}
\newcommand{\mbar}{\mbox{$\:\bar{}\:$}}
\newcommand{\mbarsp}{\mbox{$\:\bar{}\:$ }}
\newcommand{\mbarls}{\mbox{$\,\bar{}\,$}}

\newenvironment{sproof}{\noindent{\bf Proof:}\hspace*{0.2em}}{\qed\bigskip}

\newenvironment{solutionFollowedByNewParagraph}{{\bf Solution}. }{\vspace*{-0.03in} \\}

\newcommand{\bs}{\begin{solution}}
\newcommand{\es}{\end{solution}}
\newcommand{\bsnp}{\begin{solutionFollowedByNewParagraph}}
\newcommand{\esnp}{\end{solutionFollowedByNewParagraph}}

\newenvironment{alphaEnum}{\begin{list}{(\alph{enumi})}{\usecounter{enumi}
\setlength{\leftmargin}{0.44in}}}{\end{list}}
\newcommand{\mprime}[1]{\mbox{$#1'$}}

\newcommand{\fcomp}{\mbox{$\cdot$}}
\newcommand{\fcompls}{\mbox{$\:\cdot\:$}}
\newcommand{\fcompsp}{\mbox{$\cdot$} }

\newcommand{\lam}[2]{\mbox{$\lambda\,#1\,.\,#2$}}
\newcommand{\lamsp}[2]{\mbox{$\lambda\,#1\,.\,#2$} }
\newcommand{\plam}[2]{\mbox{$(\lambda\,#1\,.\,#2)$}}
\newcommand{\plamsp}[2]{\mbox{$(\lambda\,#1\,.\,#2$)} }

\newcommand{\fix}[2]{\mbox{$\mbf{fix}\,#1\,.\,#2$}}
\newcommand{\fixsp}[2]{\mbox{$\mbf{fix}\,#1\,.\,#2$} }

\newcommand{\wh}[1]{\mbox{$\widehat{#1}$}}

\newcommand{\ol}[1]{\mbox{$\overline{#1}$}}

\newcommand{\ul}[1]{\mbox{$\underline{#1}$}}

\newcommand{\nth}[1]{\mbox{$#1^{\mbox{\em th}}$}}
\newcommand{\nthsp}[1]{\mbox{$#1^{\mbox{\em th}}$} }

\newcommand{\darr}{\mbox{$\downarrow$}}
\newcommand{\uarr}{\mbox{$\downarrow$}}

\newcommand{\mless}{\mbox{\symbol{60}}}
\newcommand{\mlesssp}{\mbox{\symbol{60}} }

\newcommand{\mgreater}{\mbox{\symbol{62}}}
\newcommand{\mgreatersp}{\mbox{\symbol{62}} }

\newcommand{\types}{\mbox{$\vdash$}}
\newcommand{\typessp}{\mbox{$\vdash$} }
\newcommand{\typesls}{\mbox{$\;\vdash\;$}}

\newcommand{\mmin}[2]{\mbox{$\mbox{min}_{#1}\,#2$}}
\newcommand{\mmax}[2]{\mbox{$\mbox{max}_{#1}\,#2$}}

\newcommand{\proves}{\mbox{$\vdash$}}
\newcommand{\provesp}{\mbox{$\vdash$} }
\newcommand{\provesls}{\mbox{$\;\vdash\:$}}

\newcommand{\tproves}[1]{\mbox{$\vdash_{#1}$}}
\newcommand{\tprovesp}[1]{\mbox{$\vdash_{#1}$} }
\newcommand{\tprovesls}[1]{\mbox{$\:\vdash_{#1}\:$}}

\newcommand{\tnproves}[1]{\mbox{$\not{\hspace*{-0.007in}\vdash_{#1}}$}}
\newcommand{\tnprovesp}[1]{\mbox{$\not{\hspace*{-0.007in}\vdash_{#1}}$} }
\newcommand{\tnprovesls}[1]{\mbox{$\not{\hspace*{-0.007in}\vdash_{#1}}\:$} }

\newcommand{\mdef}[1]{\mbox{\em #1\/}}

\newcommand{\mdash}{\mbox{\symbol{45}}}
\newcommand{\mdashsp}{\mbox{\symbol{45}} }

\newcommand{\singleton}[1]{\mbox{$\langle #1 \rangle$}}

\newcommand{\mpair}[2]{\mbox{$\langle #1, #2\rangle$}}
\newcommand{\mpairsp}[2]{\mbox{$\langle #1, #2\rangle$} }

\newcommand{\mtriple}[3]{\mbox{$\langle #1, #2, #3\rangle$}}
\newcommand{\mtriplesp}[3]{\mbox{$\langle #1, #2, #3\rangle$} }

\newcommand{\mquadruple}[4]{\mbox{$\langle #1, #2, #3, #4\rangle$}}
\newcommand{\mquadruplesp}[4]{\mbox{$\langle #1, #2, #3, #4\rangle$} }

\newcommand{\us}{\symbol{95}}

\newcommand{\ttlp}{\mbox{\tt (}}
\newcommand{\sttlp}{\mbox{\small\tt (}}
\newcommand{\ttlb}{\mbox{\tt [}}
\newcommand{\sttlb}{\mbox{\small\tt [}}
\newcommand{\ttlpsp}{\mbox{\tt ( }}
\newcommand{\ttrp}{\mbox{\tt )}}
\newcommand{\sttrp}{\mbox{\small\tt )}}
\newcommand{\ttrb}{\mbox{\tt ]}}
\newcommand{\sttrb}{\mbox{\small\tt ]}}
\newcommand{\ttrpsp}{\mbox{\tt ) }}

\newcommand{\lamc}{\mbox{$\lambda$-calculus}}
\newcommand{\lamcsp}{\mbox{$\lambda$-calculus} }

\newcommand{\concat}[2]{\mbox{$#1\mbox{::}#2$}}
\newcommand{\append}[2]{\mbox{$#1\oplus#2$}}
\newcommand{\appendsymbol}{\mbox{$\oplus$}}
\newcommand{\appendsymbolsp}{\mbox{$\oplus$} }

\newcommand{\prefix}{\mbox{$\sqsubset$}}
\newcommand{\prefixeq}{\mbox{$\sqsubseteq$}}
\newcommand{\prefixls}{\mbox{$\:\sqsubset\:$}}
\newcommand{\prefixeqls}{\mbox{$\:\sqsubseteq\:$}}

\newcommand{\card}[1]{\mbox{$|#1|$}}

\newcommand{\longpage}{\enlargethispage{\baselineskip}}
\newcommand{\shortpage}{\enlargethispage{-\baselineskip}}
\newcommand{\clearemptydoublepage}{\newpage{\pagestyle{empty}\cleardoublepage}}

\newcommand{\mphi}{\mbox{$\Phi$}}
\newcommand{\mphisp}{\mbox{$\Phi$} }

\newcommand{\rsp}{\mbox{$\;\;\;\;\;$}}

\newcommand{\aquote}[1]{\mbox{$\lceil #1\rceil$}}

\newcommand{\elsesym}{\mbox{$\diamondsuit$}}
\newcommand{\elsesymls}{\mbox{$\,\diamondsuit\,$}}
\newcommand{\elsesymsp}{\mbox{$\diamondsuit$} }
\newcommand{\mcond}[3]{\mbox{$#1\,$?$\,\rarrls #2 \, \elsesym \,#3$}}

\newcommand{\derives}{\mbox{$\Vdash$}}
\newcommand{\derivesp}{\mbox{$\Vdash$} }
\newcommand{\derivesls}{\mbox{$\:\Vdash\:$}}

\newcommand{\subderives}[1]{\mbox{$\Vdash_{\mbox{\scriptsize #1}}$}}
\newcommand{\subderivesp}[1]{\mbox{$\Vdash_{\mbox{\scriptsize #1}}$} }
\newcommand{\subderivesls}[1]{\mbox{$\:\Vdash_{\mbox{\scriptsize #1}}\:$}}

\newcommand{\subprovesls}[1]{\mbox{$\:\vdash_{\mbox{\scriptsize #1}}\:$}}
\newcommand{\subproves}[1]{\mbox{$\vdash_{\mbox{\scriptsize #1}}$}}
\newcommand{\subprovesp}[1]{\mbox{$\vdash_{\mbox{\scriptsize #1}}$} }

\newcommand{\propsof}[1]{\mbox{$\mbf{Prop}[#1]$}}
\newcommand{\propsofsp}[1]{\mbox{$\mbf{Prop}[#1]$} }

\newcommand{\abof}[1]{\mbox{$\mbf{AB}[#1]$}}
\newcommand{\abofsp}[1]{\mbox{$\mbf{AB}[#1]$} }

\newcommand{\dedsof}[1]{\mbox{$\mbf{Ded}[#1]$}}
\newcommand{\dedsofsp}[1]{\mbox{$\mbf{Ded}[#1]$} }

\newcommand{\mlb}{\textnormal{[\kern-.15em[}}
\newcommand{\mrb}{\textnormal{]\kern-.15em]}}

\newcommand{\denot}[2]{\mbox{$#1\mlb#2\mrb$}}
\newcommand{\denotsp}[2]{\mbox{$#1\mlb#2\mrb$} }

\newcommand{\yields}{\mbox{$\hookrightarrow$}}
\newcommand{\yieldsls}{\mbox{$\:\hookrightarrow\:$}}
\newcommand{\yieldsp}{\mbox{$\hookrightarrow$} }

\newcommand{\msp}{\mbox{$\;\:$}}

\newcommand{\mssp}{\mbox{$\;$}}

\newcommand{\fappsp}{\mbox{$\;\,$}}

\newcommand{\ala}[1]{\mbox{\`{a} la #1}}

\newcommand{\lc}{\mbox{$\lambda$-calculus}}
\newcommand{\lcsp}{\mbox{$\lambda$-calculus} }

\newcommand{\funtype}[2]{\mbox{$#1 \rarrls #2$}}
\newcommand{\prodtype}[2]{\mbox{$#1 \times #2$}}
\newcommand{\sumtype}[2]{\mbox{$#1 + #2$}}

\newcommand{\fpslc}{\mbox{$\lambda_{\,\rarr,\,\times,\,+}$}}
\newcommand{\fpslcsp}{\mbox{$\lambda_{\,\rarr,\,\times,\,+}$} }

\newcommand{\lapp}[2]{\mbox{$#1\;#2$}}
\newcommand{\lappsp}[2]{\mbox{$#1\;#2$} }

\newcommand{\lproj}[1]{\mbox{$\pi_1(#1)$}}
\newcommand{\rproj}[1]{\mbox{$\pi_2(#1)$}}

\newcommand{\mcase}[3]{\mbox{$\mbf{case}(#1,#2,#3)$}}
\newcommand{\mcasesp}[3]{\mbox{$\mbf{case}(#1,#2,#3)$} }

\newcommand{\linj}[2]{\mbox{$\mbf{in-left}_{#1}(#2)$}}
\newcommand{\linjsp}[2]{\mbox{$\mbf{in-left}_{#1}(#2)$} }
\newcommand{\rinj}[2]{\mbox{$\mbf{in-right}_{#1}(#2)$}}
\newcommand{\rinjsp}[2]{\mbox{$\mbf{in-right}_{#1}(#2)$} }
\newcommand{\qdot}{\mbox{$\ . \ $}}
\newcommand{\bcen}{\begin{center}}
\newcommand{\ecen}{\end{center}}
\newcommand{\mtitle}[1]{#1}
\newcommand{\mvsp}{\vspace*{0.2in}}
\newcommand{\smvsp}{\vspace*{0.1in}}
\newcommand{\vsmvsp}{\vspace*{0.05in}}

\newcommand{\lessgen}{\mbox{$\preceq$}}
\newcommand{\lessgensp}{\mbox{$\preceq$ }}
\newcommand{\varment}{\mbox{$\rarr_{\text{M}}$}}
\newcommand{\varmentsp}{\mbox{$\rarr_{\text{M}}$} }
\newcommand{\varmentls}{\mbox{$\:\rarr_{\text{M}}\:$}}
\newcommand{\algmod}{\mbox{Mod}}
\newcommand{\algmodsp}{\mbox{Mod }}
\newcommand{\gterms}{\mbox{$\terms{\fsyms}{\variables}$}}
\newcommand{\gtermsp}{\mbox{$\terms{\fsyms}{\variables}$ }}
\newcommand{\natmap}{\mbox{nat}}
\newcommand{\natmapsp}{\mbox{nat }}
\newcommand{\height}{\mbox{hgt}}
\newcommand{\heightsp}{\mbox{hgt }}
\newcommand{\subcomp}{\mbox{$\circ$}}
\newcommand{\subeq}{\mbox{$\equiv$}}
\newcommand{\subeqls}{\mbox{$\:\equiv\:$}}
\newcommand{\subeqsp}{\text{$\equiv$ }}
\newcommand{\subcompsp}{\mbox{$\circ$} }
\newcommand{\subcompls}{\mbox{$\:\circ\:$}}
\newcommand{\eqsfv}{\text{$\text{Eqn}(\fsyms,V)$}}
\newcommand{\eqsfvsp}{\text{$\text{Eqn}(\fsyms,V)$} }
\newcommand{\subalg}[1]{\text{\em Sub$\{\hspace*{-0.01in}#1\}$}}
\newcommand{\tmsubalg}[1]{\text{Sub$(\hspace*{-0.01in}#1)$}} 
\newcommand{\simalg}[1]{$\text{Sim}\{#1\}$}
\newcommand{\con}{\tem{Con}}
\newcommand{\conls}{\temls{Con}}
\newcommand{\consp}{\temv{Con} }
\newcommand{\conTMode}{\temTMode{Con}}
\newcommand{\conlsTMode}{\temlsTMode{Con}}
\newcommand{\conspTMode}{\temvTMode{Con} }
\newcommand{\iso}{\text{$\cong$}}
\newcommand{\isosp}{\text{$\cong$ }}
\newcommand{\isols}{\text{$\:\cong\:$}}
\newcommand{\grar}{\text{\bf GA}}
\newcommand{\grarsp}{\text{\bf GA} }
\newcommand{\pa}{\text{\mc{P}\hspace*{-0.02in}\mc{A}}}
\newcommand{\pasp}{\text{\mc{P}\hspace*{-0.02in}\mc{A}} }
\newcommand{\quot}[1]{\text{$#1/\hspace*{-0.04in}\equiv$}}
\newcommand{\quotls}[1]{\text{$#1/\hspace*{-0.04in}\equiv\,$}}
\newcommand{\quotsp}[1]{\text{$#1/\hspace*{-0.04in}\equiv$ }}
\newcommand{\indexQuot}[2]{\text{$#1/\hspace*{-0.04in}\equiv_{#2}$}}
\newcommand{\indexQuotls}[2]{\text{$#1/\hspace*{-0.04in}\equiv_{#2}\,$}}
\newcommand{\indexQuotsp}[2]{\text{$#1/\hspace*{-0.04in}\equiv_{#2}$ }}
\newcommand{\alg}[1]{\text{$\text{Alg}(#1)$}}
\newcommand{\salg}{\text{$\mc{A}$}}
\newcommand{\salgsp}{\text{$\mc{A}$} }

\newcommand{\ranvar}[1]{\text{\rm $\text{\em RanVar\/}(#1)$}}
\newcommand{\ranvarsp}{\text{\em RanVar} }
\newcommand{\ranvarnsp}{\text{\em RanVar}}
\newcommand{\supp}[1]{\text{\rm $\text{\em Supp\/}(#1)$}}
\newcommand{\uppsp}{\text{\em Supp} }
\newcommand{\tvar}{\text{Var}}
\newcommand{\tvarsp}{\text{Var} }

\newcommand{\unifset}[1]{\text{$U(#1)$}}
\newcommand{\unifsetsp}{\text{$U$}}

\newcommand{\termdom}[1]{\text{$\temls{TDom}(#1)$}}
\newcommand{\termdomsp}{\temv{TDom} }
\newcommand{\fdom}[1]{\text{$\temls{FunDom}(#1)$}}
\newcommand{\fdomsp}{\temv{FunDom} }
\newcommand{\vdom}[1]{\text{$\temls{VarDom}(#1)$}}
\newcommand{\vdomsp}{\temv{VarDom} }

\newcommand{\comp}{\mbox{$\cdot$}}
\newcommand{\compsp}{\mbox{$\cdot$} }
\newcommand{\compls}{\mbox{$\:\cdot\:$}}

\newcommand{\ssig}{\text{$\Sigma$}}
\newcommand{\ssigsp}{\text{$\Sigma$} }

\newcommand{\termlabel}[1]{\text{$\temls{TLabel}(#1)$}}
\newcommand{\termlabelsp}{\temv{TLabel} }
\newcommand{\termlabelnsp}{\tem{TLabel}}

\newcommand{\hil}{\text{$\mathcal{H}$}}
\newcommand{\hilb}{\text{\hil\hspace*{-0.02in}}}
\newcommand{\hilsp}{\text{$\mathcal{H}$ }}
\newcommand{\hilder}{\text{$\vdash_{\mathcal{H}}$}}
\newcommand{\hilderls}{\text{$\:\vdash_{\mathcal{H}}\:$}}
\newcommand{\hildersp}{\text{$\vdash_{\mathcal{H}}$ }}
\newcommand{\jder}{\text{$\vdash_{\mathcal{J}}$}}
\newcommand{\jderls}{\text{$\:\vdash_{\mathcal{J}}\:$}}
\newcommand{\jdersp}{\text{$\vdash_{\mathcal{J}}$ }}
\newcommand{\hilj}{\text{$\mathcal{J}$}}
\newcommand{\hiljb}{\text{$\mathcal{J}$\hspace*{-0.02in}}}
\newcommand{\hiljsp}{\text{$\mathcal{J}$ }}
\newcommand{\hiljder}{\text{$\vdash_{\mathcal{J}}$}}
\newcommand{\hiljderls}{\text{$\:\vdash_{\mathcal{J}}\:$}}
\newcommand{\hiljdersp}{\text{$\vdash_{\mathcal{J}}$ }}
\newcommand{\propform}{\text{\bf P}}
\newcommand{\propformsp}{\text{\bf P} }
\newcommand{\judge}{\text{$\Gamma \rarrls A$}}
\newcommand{\judgesp}{\text{$\Gamma \rarrls A$} }
\newcommand{\nhil}[1]{\text{$\hil_{#1}$}}
\newcommand{\nhilsp}[1]{\text{$\hil_{#1}$ }}
\newcommand{\nhilj}[1]{\text{$\hilj_{#1}$}}
\newcommand{\nhiljsp}[1]{\text{$\hilj_{#1}$ }}

\newcommand{\zhil}{\text{$\hilb_0$}}
\newcommand{\zhilsp}{\text{$\hilb_0$ }}
\newcommand{\zhilj}{\text{$\hiljb_0$}}
\newcommand{\zhiljsp}{\text{$\hiljb_0$ }}
\newcommand{\ohil}{\text{$\hilb_1$}}
\newcommand{\ohilsp}{\text{$\hilb_1$ }}
\newcommand{\ohilj}{\text{$\hiljb_1$}}
\newcommand{\ohiljsp}{\text{$\hiljb_1$ }}

\newcommand{\zhilder}{\text{$\vdash_{\mathcal{H}_0}$}}
\newcommand{\zhilderls}{\text{$\:\vdash_{\mathcal{H}_0}\:$}}
\newcommand{\zhildersp}{\text{$\vdash_{\mathcal{H}_0}$ }}

\newcommand{\ohilder}{\text{$\vdash_{\mathcal{H}_1}$}}
\newcommand{\ohilderls}{\text{$\:\vdash_{\mathcal{H}_1}\:$}}
\newcommand{\ohildersp}{\text{$\vdash_{\mathcal{H}_1}$ }}

\newcommand{\zhiljder}{\text{$\vdash_{\mathcal{J}_0}$}}
\newcommand{\zhiljderls}{\text{$\:\vdash_{\mathcal{J}_0}\:$}}
\newcommand{\zhiljdersp}{\text{$\vdash_{\mathcal{J}_0}$ }}
\newcommand{\ohiljder}{\text{$\vdash_{\mathcal{J}_1}$}}
\newcommand{\ohiljderls}{\text{$\:\vdash_{\mathcal{J}_1}\:$}}
\newcommand{\ohiljdersp}{\text{$\vdash_{\mathcal{J}_1}$ }}


\newcommand{\sfola}{\text{$\Sigma_{\text{\it fol}}$}}
\newcommand{\sfolasp}{\text{$\Sigma_{\text{\it fol}}$} }
\newcommand{\lvoc}{\text{$\mathcal{L}$}}
\newcommand{\lvocsp}{\text{$\mathcal{L}$} }
\newcommand{\variables}{\text{$\mathcal{V}$}}
\newcommand{\variablesp}{\text{$\mathcal{V}$} }
\newcommand{\svars}{\text{$V$}}
\newcommand{\svarsp}{\text{$V$}}
\newcommand{\csyms}{\text{$\mathcal{C}$}}
\newcommand{\csymsp}{\text{$\mathcal{C}$} }
\newcommand{\fsyms}{\text{$\mathcal{F}$}}
\newcommand{\fsymsp}{\text{$\mathcal{F}$} }
\newcommand{\rsyms}{\text{$\mathcal{R}$}}
\newcommand{\rsymsp}{\text{$\mathcal{R}$} }
\newcommand{\psig}{\text{$\mathbf{\Omega}$}}
\newcommand{\psigvar}{\text{$(\mathbf{\Omega},\variables)$}}
\newcommand{\fsig}{\text{$\mathbf{\Sigma}$}}
\newcommand{\spsig}{\text{$(\csyms,\fsyms,\rsyms)$}}
\newcommand{\spsigsp}{\text{$(\consyms,\fsyms,\rsyms)$} }

\newcommand{\svoc}{\text{$(\psig,\variables)$}}
\newcommand{\svocsp}{\text{$(\psig,\variables)$} }

\newcommand{\psigsp}{\text{$\mathbf{\Omega}$} }
\newcommand{\consymbols}{\text{$\mathcal{C}$}}
\newcommand{\consymbolsp}{\text{$\mathcal{C}$} }
\newcommand{\funsig}{\text{$\mathcal{F}$}}
\newcommand{\funsigsp}{\text{$\mathcal{F}$} }
\newcommand{\relsig}{\text{$\mathcal{R}$}}
\newcommand{\relsigsp}{\text{$\mathcal{R}$} }

\newcommand{\arlvoc}{\text{$\mathcal{L}_{\text{\footnotesize\em ar}}$}}
\newcommand{\arlvocsp}{\text{$\mathcal{L}_{\text{\footnotesize\em ar}}$ }}

\newcommand{\arsig}{\text{$\psig_{\text{\em Ar}}$}}
\newcommand{\arsigsp}{\text{$\psig_{\text{\em Ar}}$} }

\newcommand{\lfam}{\text{$\mc{L}_{\text{\em Fam}}$}}
\newcommand{\lfamsp}{\text{$\mc{L}_{\text{\em Fam}}$ }}

\newcommand{\lblock}{\text{$\mc{L}_{\text{\em Blocks}}$}}
\newcommand{\lblocksp}{\text{$\mc{L}_{\text{\em Blocks}}$ }}

\newcommand{\inter}{\mc{I}}
\newcommand{\intersp}{\mc{I} }
\newcommand{\struc}{\mc{D}}
\newcommand{\strucsp}{\mc{D} }
\newcommand{\interar}{\text{$\mathcal{I}_{\text{\em Ar}}$}}
\newcommand{\interarsp}{\text{$\mathcal{I}_{\text{\em Ar}}$ }}

\newcommand{\arstruc}{\text{$\struc_{\text{\em Ar}}$}}
\newcommand{\arstrucsp}{\text{$\struc_{\text{\em Ar}}$} }

\newcommand{\form}[1]{\text{$\text{\bf Form}_{#1}$}}
\newcommand{\tform}[2]{\text{$\text{\bf Form}(#1,#2)$}}
\newcommand{\forml}{\form{\lvoc}}
\newcommand{\formlsp}{\form{\lvoc} }

\newcommand{\alphaeq}{\text{$\approx_{\alpha}$}}
\newcommand{\alphaeqsp}{\text{$\approx_{\alpha}$} }
\newcommand{\alphaeqls}{\text{$\:\approx_{\alpha}\,$}}

\newcommand{\aleq}{\text{$\approx_{\alpha}$}}
\newcommand{\aleqsp}{\text{$\approx_{\alpha}$} }
\newcommand{\aleqls}{\text{$\:\approx_{\alpha}\,$}}

\newcommand{\appv}[2]{\text{$#1(#2)$}}
\newcommand{\appt}[2]{\text{$\hex{#1}(#2)$}}
\newcommand{\appte}[2]{\text{$#1\,#2$}}
\newcommand{\appf}[2]{\text{$#1\,#2$}}

\newcommand{\valappt}[2]{\text{$\hex{#1}(#2)$}}
\newcommand{\valappv}[2]{\text{$#1(#2)$}}

\newcommand{\svalappt}[3]{\text{$\hex{#1}_{#2}(#3)$}}

\newcommand{\quant}[3]{\text{$(#1\,#2)\,#3$}}

\newcommand{\fvar}[1]{\text{\rm $\text{\em FV\/}(#1)$}}
\newcommand{\fvarsp}{\text{\rm\text{\em FV\/}} }

\newcommand{\bvar}[1]{\text{\rm $\text{\em BV\/}(#1)$}}
\newcommand{\bvarsp}{\text{\rm\text{\em BV\/}} }

\newcommand{\valmod}[2]{\text{\rm $#1/#2$}}

\newcommand{\svalmod}[3]{\text{\rm $#1/_{#2}\,#3$}}

\newcommand{\slvoc}{\text{$(\consymbols,\funsig,\relsig,\variables)$}}
\newcommand{\slvocsp}{\text{$(\consymbols,\funsig,\relsig,\variables)$} }

\newcommand{\ssub}[2]{\text{$\{#1 \mapsto #2\}$}}
\newcommand{\extend}[3]{\text{$#1[#2 \mapsto #3]$}}
\newcommand{\hv}{\text{$\mathbf{V}$}}
\newcommand{\hvsp}{\text{$\mathbf{V}$} }

\newcommand{\allvars}{\text{$\variables \cup \mathbf{V}$}}
\newcommand{\allvarsp}{\text{$\variables \cup \mathbf{V}$} }

\newcommand{\hlvoc}{\text{$\lvoc^{\hv}$}}
\newcommand{\hlvocsp}{\text{$\lvoc^{\hv}$} }

\newcommand{\hvoc}{\text{$(\psig,\variables \cup \hv)$}}
\newcommand{\hvocsp}{\text{$(\psig,\variables \cup \hv)$} }

\newcommand{\jone}{\text{$\mc{J}_1$}}
\newcommand{\jonesp}{\text{$\mc{J}_1$} }

\newcommand{\jprovesls}{\tprovesls{\jone}}

\newcommand{\njproves}{\text{$\not{\hspace*{-0.007in}\vdash_{\jone}$}}}

\newcommand{\njprovesp}{\text{$\not{\hspace*{-0.007in}\vdash_{\jone}}$} }

\newcommand{\njprovesls}{\text{$\:\not{\hspace*{-0.007in}\vdash_{\jone}}\,$}}

\newcommand{\mequiv}{\text{$\equiv$}}
\newcommand{\mequivls}{\text{$\:\equiv\:$}}
\newcommand{\mequivsp}{\text{$\equiv$} }

\newcommand{\cnd}{\text{\mc{NDL}}}
\newcommand{\cndsp}{\text{\mc{NDL}} }

\newcommand{\ndl}{\text{\mc{NDL}}}
\newcommand{\ndlsp}{\text{\mc{NDL}} }

\newcommand{\cndo}{\text{$\cnd_1$}}
\newcommand{\cndosp}{\text{$\cnd_1$} }

\newcommand{\cndz}{\text{$\cnd_0$}}
\newcommand{\cndzsp}{\text{$\cnd_0$} }

\newcommand{\ndlz}{\text{$\cnd_0$}}
\newcommand{\ndlzsp}{\text{$\cnd_0$} }

\newcommand{\ndlo}{\text{$\cnd_1$}}
\newcommand{\ndlosp}{\text{$\cnd_1$} }

\newcommand{\cndzomega}{\text{$\cnd_0^{\omega}$}}
\newcommand{\cndzomegasp}{\text{$\cnd_0^{\omega}$} }

\newcommand{\ndlzomega}{\text{$\cnd_0^{\omega}$}}
\newcommand{\ndlzomegasp}{\text{$\cnd_0^{\omega}$} }

\newcommand{\ndloomega}{\text{$\cnd_1^{\omega}$}}
\newcommand{\ndloomegasp}{\text{$\cnd_1^{\omega}$} }

\newcommand{\cndoomega}{\text{$\cnd_1^{\omega}$}}
\newcommand{\cndoomegasp}{\text{$\cnd_1^{\omega}$} }

\newcommand{\rds}{\mbox{${\cal R}$}}
\newcommand{\rdsp}{\mbox{${\cal R}$ }}
\newcommand{\rder}{\mbox{$\vdash_{\cal R}$}}
\newcommand{\rderls}{\mbox{$\: \vdash_{\cal R}\:$}}
\newcommand{\rdersp}{\mbox{$\vdash_{\cal R}$ }}

\newcommand{\cndprovesls}{\text{$\:\vdash_{\text{\scriptsize \cnd}}\:$}}
\newcommand{\cndproves}{\text{$\vdash_{\text{\scriptsize \cnd}}$}}
\newcommand{\cndprovesp}{\text{$\vdash_{\text{\scriptsize \cnd}}$} }
\newcommand{\ncndprovesls}
{\text{$\:\not{\hspace*{-0.007in}\vdash_{\scriptsize \cnd}}\:$}}

\newcommand{\wf}{\text{$\vdash_{\text{\em\tiny W}}\;$}}
\newcommand{\wfnsp}{\text{$\vdash_{\text{\em\scriptsize W}}$}}
\newcommand{\wfsp}{\text{$\vdash_{\text{\em\scriptsize W}}$}}

\newcommand{\man}{\text{$\mathcal{M}$}}
\newcommand{\mansp}{\text{$\mathcal{M}$ }}
\newcommand{\mander}{\text{$\vdash_{\mathcal{M}}$}}
\newcommand{\manderls}{\text{$\:\vdash_{\mathcal{M}}\:$}}
\newcommand{\mandersp}{\text{$\vdash_{\mathcal{M}}$ }}

\newcommand{\myprop}{\mbox{\bf P}}
\newcommand{\mypropsp}{\mbox{\bf P} }

\newcommand{\props}{\text{Props}}

\newcommand{\ainter}{\text{$\mathcal{I}$}}  
\newcommand{\aintersp}{\text{$\mathcal{I}$} }  
\newcommand{\aprop}{\text{\bf Prop}}
\newcommand{\apropsp}{\text{\bf Prop} }
\newcommand{\aide}{\text{\bf Ide}}
\newcommand{\aidesp}{\text{\bf Ide} }
\newcommand{\freeid}{\text{\em FId}$\,$}
\newcommand{\freeidls}{\text{\em FId$\,$ }}
\newcommand{\freeidsp}{\text{\em FId\/} }
\newcommand{\freepatid}{\text{\em FPtId}$\,$}
\newcommand{\freepatidls}{\text{\em FPtId$\,$ }}
\newcommand{\freepatidsp}{\text{\em FPtId\/} }
\newcommand{\patvar}{\text{\em PtVar}$\,$}
\newcommand{\patvarsp}{\text{\em PtVar$\,$} }
\newcommand{\patidsp}{\text{\em PatId\/} }
\newcommand{\avar}{\text{\bf Var}}
\newcommand{\avarsp}{\text{\bf Var} }
\newcommand{\apat}{\text{\bf Pat}}
\newcommand{\apatsp}{\text{\bf Pat} }
\newcommand{\compat}{\text{\bf ComPat}}
\newcommand{\compatsp}{\text{\bf ComPat} }
\newcommand{\aexp}{\text{\bf Exp}}
\newcommand{\aexpsp}{\text{\bf Exp} }
\newcommand{\aded}{\text{\bf Ded}}
\newcommand{\adedsp}{\text{\bf Ded} }
\newcommand{\aphrase}{\text{\bf Phr}}
\newcommand{\aphrasesp}{\text{\bf Phr} }
\newcommand{\avexp}{\text{\bf ValExp}}
\newcommand{\avexpsp}{\text{\bf ValExp} }
\newcommand{\ttrue}{\text{\tt true}}
\newcommand{\ttruesp}{\text{\tt true} }
\newcommand{\ttfalse}{\text{\tt false}}
\newcommand{\ttfalsesp}{\text{\tt false} }
\newcommand{\aerror}{\text{\em error\/}}
\newcommand{\aerrorsp}{\text{\em error\/ }}
\newcommand{\idwo}{\text{$\prec$}}
\newcommand{\idwosp}{\text{$\prec$} }
\newcommand{\adom}{\text{$\mathcal{D}$}}
\newcommand{\adomsp}{\text{$\mathcal{D}$ }}
\newcommand{\aval}{\text{\bf Val}}
\newcommand{\avalsp}{\text{\bf Val }}
\newcommand{\propval}{\text{\bf PropVal}}
\newcommand{\propvalsp}{\text{\bf PropVal }}
\newcommand{\fval}{\text{\bf FVal}}
\newcommand{\fvalsp}{\text{\bf FVal }}
\newcommand{\tval}{\text{\bf MVal}}
\newcommand{\tvalsp}{\text{\bf MVal }}
\newcommand{\topval}{\text{\bf TopVal}}
\newcommand{\topvalsp}{\text{\bf TopVal }}
\newcommand{\propc}{\text{\em prop}}
\newcommand{\propcls}{\text{\em prop$\,$}}
\newcommand{\propcsp}{\text{\em prop\/} }
\newcommand{\fclos}{\text{\em fclos}}
\newcommand{\fclosls}{\text{\em fclos$\,$}}
\newcommand{\fclosp}{\text{\em fclos\/} }
\newcommand{\tclos}{\text{\em mclos}}
\newcommand{\tclosls}{\text{\em mclos$\,$}}
\newcommand{\tclosp}{\text{\em mclos\/} }
\newcommand{\aop}{\text{\em op}}
\newcommand{\aopls}{\text{\em op$\,$}}
\newcommand{\aopsp}{\text{\em op} }
\newcommand{\arule}{\text{\em rule}}
\newcommand{\arulels}{\text{\em rule$\,$}}
\newcommand{\arulesp}{\text{\em rule} }
\newcommand{\aenv}{\text{$\rho$}}
\newcommand{\aenvsp}{\text{$\rho$ }}
\newcommand{\ab}{\text{$\beta$}}
\newcommand{\absp}{\text{$\beta$ }}
\newcommand{\aenvdom}{\text{\em Env}}
\newcommand{\aenvdomsp}{\text{\em Env} }
\newcommand{\aprocdom}{\text{\em Proc}}
\newcommand{\aprocdomsp}{\text{\em Proc} }
\newcommand{\unbound}{\text{\em unbound}}
\newcommand{\unboundsp}{\text{\em unbound} }
\newcommand{\emean}{\text{$\mathcal{E}$}}
\newcommand{\emeansp}{\text{$\mathcal{E}$ }}
\newcommand{\cmean}{\text{$\mathcal{A}$}}
\newcommand{\cmeansp}{\text{$\mathcal{A}$ }}
\newcommand{\mmean}{\text{$\mathcal{M}$}}
\newcommand{\mmeansp}{\text{$\mathcal{M}$ }}
\newcommand{\smallsome}{\text{\small SOME}}
\newcommand{\smallsomesp}{\text{\small SOME} }
\newcommand{\smallnone}{\text{\small NONE}}
\newcommand{\smallnonesp}{\text{\small NONE} }
\newcommand{\enew}{\text{$E_{\text{new}}$}}
\newcommand{\enewsp}{\text{$E_{\text{new}}$} }
\newcommand{\isub}{\text{$I_{\text{sub}}$}}
\newcommand{\isubsp}{\text{$I_{\text{sub}}$} }
\newcommand{\sab}{\text{$\{\enew/\isub\}$}}
\newcommand{\sabsp}{\text{$\{\enew/\isub\}$} }
\newcommand{\apsem}{\text{$\mathcal{A}\mathcal{S}$}}
\newcommand{\apsemsp}{\text{$\mathcal{A}\mathcal{S}$} }
\newcommand{\tupn}{\text{\em tup$_n$}}
\newcommand{\tupnsp}{\text{\em tup$_n$} }
\newcommand{\proj}{\text{\em proj}}
\newcommand{\projsp}{\text{\em proj} }
\newcommand{\cmode}{\text{$\mathcal{C}$}}
\newcommand{\cmodesp}{\text{$\mathcal{C}$} }
\newcommand{\dmode}{\text{$\mathcal{D}$}}
\newcommand{\dmodesp}{\text{$\mathcal{D}$} }
\newcommand{\cder}{\text{$\vdash_{\mathcal C}$}}
\newcommand{\cderls}{\text{$\;\,\vdash_{\mathcal C}\:$}}
\newcommand{\cdersp}{\text{$\vdash_{\mathcal C}$} }
\newcommand{\dder}{\text{$\vdash_{\mathcal D}$}}
\newcommand{\dderls}{\text{$\:\,\vdash_{\mathcal D}\:$}}
\newcommand{\ddersp}{\text{$\vdash_{\mathcal D}$} }
\newcommand{\ttand}{\text{\tt and}}
\newcommand{\ttandsp}{\text{\tt and }}
\newcommand{\ttor}{\text{\tt or}}
\newcommand{\ttorsp}{\text{\tt or }}
\newcommand{\ttif}{\text{\tt if}}
\newcommand{\ttifsp}{\text{\tt if }}
\newcommand{\ttiff}{\text{\tt iff}}
\newcommand{\ttiffsp}{\text{\tt iff }}
\newcommand{\ttnot}{\text{\tt not}}
\newcommand{\ttnotsp}{\text{\tt not }}
\newcommand{\ttqm}{\text{\tt ?}}
\newcommand{\ttqmsp}{\text{\tt ?} }
\newcommand{\ttbq}{\text{\tt `}}
\newcommand{\ttbqsp}{\text{\tt `} }
\newcommand{\ttdollar}{\text{\tt \$}}
\newcommand{\ttdollarsp}{\text{\tt \$} }

\newcommand{\ttin}{\text{\tt in}}
\newcommand{\ttinsp}{\text{\tt in }}
\newcommand{\ttul}{\text{\tt \_}}
\newcommand{\ttulsp}{\text{\tt \_ }}
\newcommand{\topname}{\text{\em TopName}}
\newcommand{\topnamesp}{\text{\em TopName\/ }}
\newcommand{\logop}{\text{\em LogOp}}
\newcommand{\logopsp}{\text{\em LogOp\/ }}
\newcommand{\primrule}{\text{\em Prim-Rule}}
\newcommand{\primrulesp}{\text{\em Prim-Rule\/ }}
\newcommand{\mlceil}{\text{$\lceil$}}
\newcommand{\mlceilsp}{\text{$\lceil$ }}
\newcommand{\mrceil}{\text{$\rceil$}}
\newcommand{\mrceilsp}{\text{$\rceil$ }}
\newcommand{\mlfloor}{\text{$\lfloor$}}
\newcommand{\mlfloorsp}{\text{$\lfloor$ }}
\newcommand{\mrfloor}{\text{$\rfloor$}}
\newcommand{\mrfloorsp}{\text{$\rfloor$ }}
\newcommand{\eval}[1]{\text{$\temls{Eval}(#1)$}}
\newcommand{\evalsp}{\text{\em Eval\/} }
\newcommand{\evalnsp}{\temv{Eval}}
\newcommand{\sev}[1]{\text{$\temls{ev}(#1)$}}
\newcommand{\evals}{\text{$\:\leadsto\;$}}
\newcommand{\tevals}{\text{$\:\leadsto^*\:$}}
\newcommand{\eevals}{\text{$\:\hookrightarrow\:$}}
\newcommand{\error}{\text{\tem{error}}}
\newcommand{\errorsp}{\text{\tem{error}} }
\newcommand{\proptype}{\text{\bf prop}}
\newcommand{\proptypesp}{\text{\bf prop }}
\newcommand{\env}{\text{\bf Env}}
\newcommand{\envsp}{\text{\bf Env }}

\newcommand{\fdl}{\text{\mc{FND}}}
\newcommand{\fdlsp}{\text{\mc{FND}} }

\newcommand{\obeqproves}{\text{$\vdash_{\text{\scriptsize OE}}\,$}}

\newcommand{\obeq}{\text{$\approx$}}
\newcommand{\obeqsp}{\text{$\approx$} }
\newcommand{\obeqls}{\text{$\:\approx\,$}}
\newcommand{\nobeqls}{\text{$\:\not\approx\,$}}

\newcommand{\subobeq}[1]{\text{$\approx_{#1}$}}
\newcommand{\subobeqsp}[1]{\text{$\approx_{#1}$} }
\newcommand{\subobeqls}[1]{\text{$\:\approx_{#1}\,$}}
\newcommand{\subnobeqls}[1]{\text{$\:\not\approx_{#1}\,$}}

\newcommand{\give}{\text{$\rightarrowtail$}}
\newcommand{\givesp}{\text{$\rightarrowtail$} }
\newcommand{\gives}{\text{$\,\rightarrowtail\:$}}
\newcommand{\ngives}{\text{$\:\not\rightarrowtail\,$}}

\newcommand{\dep}{\text{$\twoheadrightarrow$}}
\newcommand{\depsp}{\text{$\twoheadrightarrow$} }
\newcommand{\depls}{\text{$\,\twoheadrightarrow\:$}}
\newcommand{\ndepls}{\text{$\,\not\twoheadrightarrow\:$}}

\newcommand{\mdep}{\text{$\Rrightarrow$}}
\newcommand{\mdepsp}{\text{$\Rrightarrow$} }
\newcommand{\mdepls}{\text{$\,\Rrightarrow\:$}}
\newcommand{\mndepls}{\text{$\,\not\Rrightarrow\:$}}

\newcommand{\tcdep}{\text{$\twoheadrightarrow^+$}}
\newcommand{\tcdepsp}{\text{$\twoheadrightarrow^+$} }
\newcommand{\tcdepls}{\text{$\,\twoheadrightarrow^+\:$}}
\newcommand{\tcndepls}{\text{$\,\not\twoheadrightarrow^+\:$}}

\newcommand{\ttcdep}[1]{\text{$\twoheadrightarrow_{#1}^+$}}
\newcommand{\ttcdepsp}[1]{\text{$\twoheadrightarrow_{#1}^+$} }
\newcommand{\ttcdepls}[1]{\text{$\,\twoheadrightarrow_{#1}^+\:$}}
\newcommand{\ttncdepls}[1]{\text{$\,\not\twoheadrightarrow_{#1}^+\:$}}

\newcommand{\strict}{\text{$\vdash_{\text{\em Strict}}$}}
\newcommand{\strictsp}{\text{$\vdash_{\text{\em Strict}}$} }
\newcommand{\strictls}{\text{$\vdash_{\text{\em Strict}}\,$}}

\newcommand{\el}[1]{\text{$\mc{E}(#1)$}}

\newcommand{\fass}[1]{\text{\rm $\text{\em FA\/}(#1)$}}
\newcommand{\fassp}{\text{\rm\text{\em FA\/}} }
\newcommand{\fassnsp}{\text{\rm\text{\em FA}}}

\newcommand{\st}[1]{\fass{#1}}

\newcommand{\fst}[1]{\text{$\tem{Strict}(#1)$}}

\newcommand{\concl}[1]{\text{$\mc{C}(#1)$}}

\newcommand{\conclnsp}{\mc{C}}

\newcommand{\conclsp}{\mc{C} }

\newcommand{\fconcl}[1]{\text{$\tem{Con}(#1)$}}

\newcommand{\mhex}[1]{\widehat{#1}}

\newcommand{\grab}[1]{\text{$\mless #1 \mgreater\:$}}
\newcommand{\grabsp}[1]{\text{$\mless #1 \mgreater$} }

\newcommand{\sz}[1]{\text{$\temls{SZ}(#1)$}}

\newcommand{\lsz}[1]{\text{$\temls{LSZ}(#1)$}}

\newcommand{\allprops}[2]{\text{$\text{\bf Prop}(#1,#2)$}}

\newcommand{\appdsub}[3]{\text{$#1\{#2/#3\}$}}

\newcommand{\dsub}[2]{\text{$\{#1/#2\}$}}

\newcommand{\mappt}[2]{\text{$\hex{#1}(#2)$}}

\newcommand{\fa}[1]{\text{$\tem{FA}(#1)$}}

\newcommand{\abdom}{\mc{B}}
\newcommand{\abdomsp}{\mc{B} }

\newcommand{\err}{\tem{error}}
\newcommand{\errsp}{\tem{error} }

\newcommand{\PrimDed}{\mbf{PrimDed}}
\newcommand{\PrimDedsp}{\mbf{PrimDed} }

\newcommand{\pdom}{\mbf{Prop}}
\newcommand{\pdomsp}{\mbf{Prop} }

\newcommand{\dm}{\mc{D}}
\newcommand{\dmsp}{\mc{D} }
\newcommand{\pdm}{\mc{M}}
\newcommand{\pdmsp}{\mc{M} }

\newcommand{\Luk}{\mbox{{\L}ukasiewicz}}
\newcommand{\Jask}{\mbox{J\'{a}skowski}}

\newcommand{\Luksp}{\mbox{{\L}ukasiewics} }
\newcommand{\Jasksp}{\mbox{J\'{a}skowski} }

\newcommand{\ass}[2]{\text{$\mbf{assume}\;#1\,.\,#2$}}
\newcommand{\assp}[2]{\text{$\mbf{assume}\;#1\,.\,#2$} }

\newcommand{\assin}[2]{\text{$\mbf{assume}\,\;#1\,\;\mbf{in}\;#2$}}
\newcommand{\assinsp}[2]{\text{$\mbf{assume}\;#1\,\mbf{in}\;#2$} }

\newcommand{\supab}[2]{\text{$\mbf{suppose-absurd}\,\;#1\,\;\mbf{in}\;#2$}}
\newcommand{\supabsp}[2]{\text{$\mbf{suppose-absurd}\;#1\,\mbf{in}\;#2$} }

\newcommand{\abst}[2]{\text{$\lambda\;#1\,.\,#2$}}
\newcommand{\abstsp}[2]{\text{$\lambda\;#1\,.\,#2$} }

\newcommand{\type}{\text{$\tau$}}
\newcommand{\typesp}{\text{$\tau$} }

\newcommand{\xt}{\text{$x_{\tau}$}}
\newcommand{\xtsp}{\text{$x_{\tau}$} }

\newcommand{\xs}{\text{$x_{\sigma}$}}
\newcommand{\xsp}{\text{$x_{\sigma}$} }

\newcommand{\modus}[2]{\text{$\mbf{mp}(#1,#2)$}}
\newcommand{\modusp}[2]{\text{$\mbf{mp}(#1,#2)$} }

\newcommand{\app}[2]{\text{$\mbf{app}(#1,#2)$}}
\newcommand{\appsp}[2]{\text{$\mbf{app}(#1,#2)$} }

\newcommand{\both}[2]{\text{$\mbf{both}(#1,#2)$}}
\newcommand{\bothsp}[2]{\text{$\mbf{both}(#1,#2)$} }

\newcommand{\pair}[2]{\text{$\mbf{pair}(#1,#2)$}}
\newcommand{\pairsp}[2]{\text{$\mbf{pair}(#1,#2)$} }

\newcommand{\leftand}[1]{\text{$\mbf{left}(#1)$}}
\newcommand{\leftandsp}[1]{\text{$\mbf{left}(#1)$} }

\newcommand{\first}[1]{\text{$\mbf{first}(#1)$}}
\newcommand{\firstsp}[1]{\text{$\mbf{first}(#1)$} }

\newcommand{\second}[1]{\text{$\mbf{second}(#1)$}}
\newcommand{\secondsp}[1]{\text{$\mbf{second}(#1)$} }

\newcommand{\rightand}[1]{\text{$\mbf{right}(#1)$}}
\newcommand{\rightandsp}[1]{\text{$\mbf{right}(#1)$} }

\newcommand{\fv}[1]{\text{$\tem{FV}(#1)$}}
\newcommand{\fvsp}[1]{\text{$\tem{FV}(#1)$} }

\newcommand{\ch}{\text{Curry-Howard}}
\newcommand{\chsp}{\text{Curry-Howard} }

\newcommand{\genproof}{\text{$\mathfrak{D}$}}

\newcommand{\genproofsp}{\text{$\mathfrak{D}$} }

\newcommand{\paleq}{\text{$\equiv_{\epsilon}$}}
\newcommand{\paleqsp}{\text{$\equiv_{\epsilon}$} }
\newcommand{\paleqls}{\text{$\:\equiv_{\epsilon}\,$}}

\newcommand{\eigenrename}[1]{\text{$\temls{EigenRename}(#1)$}}
\newcommand{\eigenrenamensp}{\tem{EigenRename}}
\newcommand{\eigenrenamesp}{\temv{EigenRename} }

\newcommand{\evar}[1]{\text{\rm $\text{\em EV\/}(#1)$}}
\newcommand{\evarsp}{\text{\rm\text{\em EV\/}} }

\newcommand{\formdom}[1]{\text{$\temls{FDom}(#1)$}}
\newcommand{\formdomsp}{\temv{FDom} }
\newcommand{\formdomnsp}{\tem{FDom}}

\newcommand{\formlabel}[1]{\text{$\temls{FLabel}(#1)$}}
\newcommand{\formlabelsp}{\temv{FLabel} }
\newcommand{\formlabelnsp}{\tem{FLabel}}

\newcommand{\proofdom}[1]{\text{$\temls{DDom}(#1)$}}
\newcommand{\proofdomsp}{\temv{DDom} }

\newcommand{\prooflabel}[1]{\text{$\temls{DLabel}(#1)$}}
\newcommand{\prooflabelsp}{\temv{DLabel} }
\newcommand{\prooflabelnsp}{\tem{DLabel}}

\newcommand{\appd}[2]{\text{$#1\,#2$}}
\newcommand{\apppd}[2]{\text{$#1\;#2$}}

\newcommand{\deductionset}[1]{\text{$\mbf{Ded}(#1)$}}

\newcommand{\formalappf}[2]{\text{${#1}^{\sharp}(#2)$}}
\newcommand{\formalappfnsp}[1]{\text{${#1}^{\sharp}$}}
\newcommand{\formalappd}[2]{\text{${#1}^{\star}(#2)$}}
\newcommand{\formalappdnsp}[1]{\text{${#1}^{\star}$}}

\newcommand{\Phisp}{\text{$\Phi$ }}

\newcommand{\apprule}[2]{\text{$#1\; #2$}}

\newcommand{\rulemp}{\mbf{modus-ponens}}
\newcommand{\rulempsp}{\mbf{modus-ponens} }
\newcommand{\ruleclaim}{\mbf{claim}}
\newcommand{\ruleclaimsp}{\mbf{claim} }
\newcommand{\rulemt}{\mbf{modus-tollens}}
\newcommand{\rulemtsp}{\mbf{modus-tollens} }
\newcommand{\ruledn}{\mbf{double-negation}}
\newcommand{\rulednsp}{\mbf{double-negation} }
\newcommand{\ruleland}{\mbf{left-and}}
\newcommand{\rulelandsp}{\mbf{left-and} }
\newcommand{\rulerand}{\mbf{right-and}}
\newcommand{\rulerandsp}{\mbf{right-and} }

\newcommand{\ruleliff}{\mbf{left-iff}}
\newcommand{\ruleriff}{\mbf{right-iff}}
\newcommand{\ruleequiv}{\mbf{equivalence}}

\newcommand{\ruleboth}{\mbf{both}}
\newcommand{\rulebothsp}{\mbf{both} }

\newcommand{\ruleabsurd}{\mbf{absurd}}
\newcommand{\ruleabsurdsp}{\mbf{absurd} }

\newcommand{\zlevel}{\text{0-level}}
\newcommand{\zlevelsp}{\text{0-level} }
\newcommand{\Zlevel}{\text{0-level}}
\newcommand{\Zlevelsp}{\text{0-level} }

\newcommand{\typealpha}{\text{type-$\alpha$}}
\newcommand{\typealphasp}{\text{type-$\alpha$} }
\newcommand{\Typealpha}{\text{Type-$\alpha$}}
\newcommand{\Typealphasp}{\text{Type-$\alpha$} }

\newcommand{\typeomega}{\text{type-$\omega$}}
\newcommand{\typeomegasp}{\text{type-$\omega$} }
\newcommand{\Typeomega}{\text{Type-$\omega$}}
\newcommand{\Typeomegasp}{\text{Type-$\omega$} }

\newcommand{\inflevel}{\text{$\infty$-level}}
\newcommand{\inflevelsp}{\text{$\infty$-level} }
\newcommand{\Inflevel}{\text{$\infty$-level}}
\newcommand{\Inflevelsp}{\text{$\infty$-level} }

\newcommand{\mtype}[1]{\text{type-$#1$}}
\newcommand{\mtypesp}[1]{\text{type-$#1$}}

\newcommand{\mType}[1]{\text{Type-$#1$}}
\newcommand{\mTypesp}[1]{\text{Type-$#1$}}

\newcommand{\assume}[2]{\text{$\mbf{assume } #1 \mbf{ in } #2$}}
\newcommand{\msupab}[2]{\text{$\mbf{suppose-absurd } #1 \mbf{ in } #2$}}

\newcommand{\bpropcon}{\text{$\odot$}}
\newcommand{\bpropconls}{\text{$\:\odot\:$}}
\newcommand{\bpropconsp}{\text{$\odot$} }

\newcommand{\lm}{\text{$\lambda\hspace*{0.01in}\phi$}}
\newcommand{\lmsp}{\text{$\lambda\hspace*{0.01in}\phi$} }
\newcommand{\lmc}{\text{$\lambda\hspace*{0.01in}\phi$-calculus}}
\newcommand{\lmcsp}{\text{$\lambda\hspace*{0.01in}\phi$-calculus} }

\newcommand{\Phr}{\rm \text{\em Phr}}
\newcommand{\Phrsp}{\rm \text{\em Phr\/} }
\newcommand{\Exp}{\rm \text{\em Exp}}
\newcommand{\Expsp}{\rm \text{\em Exp\/} }
\newcommand{\Ded}{\rm \text{\em Ded}}
\newcommand{\Dedsp}{\rm \text{\em Ded\/} }
\newcommand{\Ide}{\rm \text{\em Ide}}
\newcommand{\Idesp}{\rm \text{\em Ide\/} }

\newcommand{\mab}[2]{\text{$\phi\,#1\,.\,#2$}}
\newcommand{\mabop}{\text{$\phi$}}
\newcommand{\mabopsp}{\text{$\phi$} }
\newcommand{\lab}[2]{\text{$\lambda\,#1\,.\,#2$}}

\newcommand{\dapp}[2]{\text{$\mbf{dapp}(#1,#2)$}}
\newcommand{\dappthunk}[1]{\text{$\mbf{dapp}(#1)$}}
\newcommand{\eapp}[2]{\text{$\mbf{app}(#1,#2)$}}
\newcommand{\eappthunk}[1]{\text{$\mbf{app}(#1)$}}

\newcommand{\dcase}[3]{\text{$\mbf{dcase}(#1,#2,#3)$}}
\newcommand{\ecase}[3]{\text{$\mbf{case}(#1,#2,#3)$}}

\newcommand{\prop}{\tem{Prop}}
\newcommand{\propsp}{\tem{Prop} }

\newcommand{\fexp}{\tem{FExp}}
\newcommand{\fexpsp}{\tem{FExp} }
\newcommand{\mexp}{\tem{MExp}}
\newcommand{\mexpsp}{\tem{MExp} }

\newcommand{\values}{\mc{V}}
\newcommand{\valuesp}{\mc{V} }

\newcommand{\conv}{\text{$\doteq$}}
\newcommand{\convls}{\text{$\:\doteq\:$}}
\newcommand{\convsp}{\text{$\doteq$} }

\newcommand{\semeq}{\text{$\approx$}}
\newcommand{\semeqls}{\text{$\:\approx\:$}}
\newcommand{\semeqsp}{\text{$\approx$} }

\newcommand{\valeq}{\text{$\equiv$}}
\newcommand{\valeqls}{\text{$\:\equiv\:$}}
\newcommand{\valeqsp}{\text{$\equiv$} }

\newcommand{\domega}{\text{$\Omega_{\mbf{D}}$}}
\newcommand{\domegasp}{\text{$\Omega_{\mbf{D}}$} }

\newcommand{\eomega}{\text{$\Omega_{\mbf{E}}$}}
\newcommand{\eomegasp}{\text{$\Omega_{\mbf{E}}$} }

\newcommand{\mvec}[1]{\text{$\overrightarrow{#1}$}}
\newcommand{\mvecsp}[1]{\text{$\overrightarrow{#1}$} }

\newcommand{\lmcon}[1]{\text{$\mathsf{#1}$}}

\newcommand{\tred}{\text{$\rightarrowtail^*$}}
\newcommand{\tredls}{\text{$\:\rightarrowtail^*\:$}}
\newcommand{\tredsp}{\text{$\rightarrowtail^*$} }

\newcommand{\convg}[2]{\text{$#1\hspace*{-0.04in}\downarrow_{#2}$}}
\newcommand{\divg}[2]{\text{$#1\hspace*{-0.04in}\uparrow_{#2}$}}

\newcommand{\sconvg}[1]{\text{$#1\hspace*{-0.04in}\downarrow$}}
\newcommand{\sdivg}[1]{\text{$#1\hspace*{-0.04in}\uparrow$}}

\newcommand{\cclaim}{\lmcon{claim}}
\newcommand{\cclaimsp}{\lmcon{claim} }

\newcommand{\cmp}{\lmcon{mp}}
\newcommand{\cmpsp}{\lmcon{mp} }

\newcommand{\ctrue}{\lmcon{true}}
\newcommand{\ctruesp}{\lmcon{true} }

\newcommand{\cfalse}{\lmcon{false}}
\newcommand{\cfalsesp}{\lmcon{false} }

\newcommand{\cmt}{\lmcon{mt}}
\newcommand{\cmtsp}{\lmcon{mt} }

\newcommand{\ctrueaxiom}{\lmcon{T\text{-}axiom}}
\newcommand{\ctrueaxiomsp}{\lmcon{T\text{-}axiom} }
\newcommand{\cfalseaxiom}{\lmcon{F\text{-}axiom}}
\newcommand{\cfalseaxiomsp}{\lmcon{F\text{-}axiom} }

\newcommand{\cdn}{\lmcon{dn}}
\newcommand{\cdnsp}{\lmcon{dn} }
\newcommand{\ccd}{\lmcon{cd}}
\newcommand{\cboth}{\lmcon{both}}
\newcommand{\cleither}{\lmcon{left\text{-}either}}
\newcommand{\creither}{\lmcon{right\text{-}either}}
\newcommand{\cequiv}{\lmcon{equiv}}
\newcommand{\cliff}{\lmcon{left\text{-}iff}}
\newcommand{\criff}{\lmcon{right\text{-}iff}}
\newcommand{\cabsurd}{\lmcon{absurd}}
\newcommand{\cland}{\lmcon{left\text{-}and}}
\newcommand{\crand}{\lmcon{right\text{-}and}}

\newcommand{\map}[1]{\text{$\mlb #1\mrb$}}
\newcommand{\mapsp}[1]{\text{$\mlb #1\mrb$} }

\newcommand{\exm}{\mbf{!}}
\newcommand{\exmsp}{\mbf{!} }
\newcommand{\exmls}{\mbf{!}\hspace*{0.01in}}

\newcommand{\noeval}{\text{$\tem{Eval}_{\text{\scriptsize\em N}}$}}

\newcommand{\aoeval}{\text{$\tem{Eval}_{\text{\scriptsize\em A}}$}}

\newcommand{\ceval}{\text{$\tem{Eval}_{\text{\scriptsize\em C}}$}}

\newcommand{\lmnot}{\lmcon{not}}
\newcommand{\lmnotsp}{\lmcon{not} }

\newcommand{\lmand}{\lmcon{and}}
\newcommand{\lmandsp}{\lmcon{and} }

\newcommand{\lmor}{\lmcon{or}}
\newcommand{\lmorsp}{\lmcon{or} }

\newcommand{\lmif}{\lmcon{if}}
\newcommand{\lmifsp}{\lmcon{if} }

\newcommand{\lmiff}{\lmcon{iff}}
\newcommand{\lmiffsp}{\lmcon{iff} }

\newcommand{\lmcond}{\mbf{cond}}
\newcommand{\lmcondsp}{\mbf{cond} }

\newcommand{\lmdcond}{\mbf{dcond}}
\newcommand{\lmdcondsp}{\mbf{dcond} }

\newcommand{\suapp}[2]{\text{$(#1\msp #2)$}}
\newcommand{\sbapp}[3]{\text{$(#1\msp #2\msp #3)$}}
\newcommand{\stapp}[4]{\text{$(#1\msp #2\msp #3\msp #4)$}}

\newcommand{\npsuapp}[2]{\text{$#1\msp #2$}}
\newcommand{\npsbapp}[3]{\text{$#1\msp #2\msp #3$}}
\newcommand{\npstapp}[4]{\text{$#1\msp #2\msp #3\msp #4$}}

\newcommand{\lmeq}{\lmcon{=}}
\newcommand{\lmeqsp}{\lmcon{=} }

\newcommand{\clp}{\lmcon{(}}
\newcommand{\clpsp}{\lmcon{(} }
\newcommand{\crp}{\lmcon{)}}
\newcommand{\crpsp}{\lmcon{)} }

\newcommand{\npumapp}[2]{\text{$\exmls #1 \msp #2$}}
\newcommand{\npbmapp}[3]{\text{$\exmls #1 \msp #2 \msp #3$}}
\newcommand{\nptmapp}[4]{\text{$\exmls #1 \msp #2 \msp #3 \msp #4$}}

\newcommand{\mthunkapp}[1]{\text{$(\exmls #1)$}}
\newcommand{\npmthunkapp}[1]{\text{$\exmls #1$}}
\newcommand{\umapp}[2]{\text{$(\exmls #1 \msp #2)$}}
\newcommand{\bmapp}[3]{\text{$(\exmls #1 \msp #2 \msp #3)$}}
\newcommand{\tmapp}[4]{\text{$(\exmls #1 \msp #2 \msp #3 \msp #4)$}}

\newcommand{\checke}{\text{$\lmcon{if}_E$}}
\newcommand{\checkesp}{\text{$\lmcon{if}_E$} }

\newcommand{\checkd}{\text{$\lmcon{if}_D$}}
\newcommand{\checkdsp}{\text{$\lmcon{if}_D$} }

\newcommand{\pdes}{\mc{T}}
\newcommand{\pdesp}{\mc{T} }

\newcommand{\booland}{\lmcon{bool\mdash and}}
\newcommand{\boolandsp}{\lmcon{bool\mdash and} }

\newcommand{\boolor}{\lmcon{bool\mdash or}}
\newcommand{\boolorsp}{\lmcon{bool\mdash or} }

\newcommand{\boolnot}{\lmcon{bool\mdash not}}
\newcommand{\boolnotsp}{\lmcon{bool\mdash not} }

\newcommand{\sequent}[2]{\text{$\langle #1,#2\rangle$}}

\newcommand{\lmzero}{\lmcon{0}}
\newcommand{\lmzerosp}{\lmcon{0} }


\newcommand{\lmsucc}[1]{\lmcon{s(}#1\lmcon{)}}
\newcommand{\lmfact}[1]{\lmcon{fact(}#1\lmcon{)}}

\newcommand{\lmplus}[2]{\lmcon{(}#1\mssp\lmcon{+}\mssp#2\lmcon{)}}

\newcommand{\nplmplus}[2]{#1\mssp\lmcon{+}\mssp#2}

\newcommand{\nplmtimes}[2]{#1\mssp\lmcon{*}\mssp#2}

\newcommand{\lmtimes}[2]{\lmcon{(}#1\mssp\lmcon{*}\mssp#2\lmcon{)}}

\newcommand{\plusri}{\lmcon{plus\mdash 1}}
\newcommand{\plusrii}{\lmcon{plus\mdash 2}}

\newcommand{\scong}{\lmcon{s\mdash cong}}
\newcommand{\scongsp}{\lmcon{s\mdash cong} }
\newcommand{\factcong}{\lmcon{fact\mdash cong}}
\newcommand{\factcongsp}{\lmcon{fact\mdash cong} }
\newcommand{\pluscong}{\lmcon{+\mdash cong}}
\newcommand{\pluscongsp}{\lmcon{+\mdash cong} }
\newcommand{\timescong}{\lmcon{*\mdash cong}}
\newcommand{\timescongsp}{\lmcon{*\mdash cong} }

\newcommand{\plusrisp}{\lmcon{plus\mdash 1} }
\newcommand{\plusriisp}{\lmcon{plus\mdash 2} }

\newcommand{\timesri}{\lmcon{times\mdash 1}}
\newcommand{\timesrii}{\lmcon{times\mdash 2}}

\newcommand{\factri}{\lmcon{fact\mdash 1}}
\newcommand{\factrii}{\lmcon{fact\mdash 2}}
\newcommand{\factriisp}{\lmcon{fact\mdash 2} }

\newcommand{\vmap}[2]{\text{$#1\mlb #2\mrb$}}
\newcommand{\vmapsp}[2]{\text{$#1\mlb #2\mrb$} }

\newcommand{\tarsk}{\text{$\succcurlyeq$}}
\newcommand{\tarskls}{\text{$\:\succcurlyeq\:$}}
\newcommand{\tarsksp}{\text{$\succcurlyeq$} }

\newcommand{\zeroeven}{\lmcon{zero\mdash axiom}}
\newcommand{\zeroaxiom}{\lmcon{zero\mdash axiom}}
\newcommand{\zeroaxiomsp}{\lmcon{zero\mdash axiom} }
\newcommand{\mkeven}{\lmcon{make\mdash even}}
\newcommand{\mkevensp}{\lmcon{make\mdash even} }
\newcommand{\mkodd}{\lmcon{make\mdash odd}}
\newcommand{\mkoddsp}{\lmcon{make\mdash odd} }
\newcommand{\podd}{\lmcon{Odd}}
\newcommand{\peven}{\lmcon{Even}}
\newcommand{\mo}{\lmcon{mo}}
\newcommand{\me}{\lmcon{me}}
\newcommand{\za}{\lmcon{za}}

\newcommand{\teq}{\text{$\approx$}}
\newcommand{\teqsp}{\text{$\approx$} }
\newcommand{\teqls}{\text{$\:\approx\:$}}

\newcommand{\uniflm}{\tem{Unif}}
\newcommand{\uniflmsp}{\temv{Unif} }

\newcommand{\unifcalc}{\mc{U}}
\newcommand{\unifcalcsp}{\mc{U} }

\newcommand{\unif}{\text{$\vdash_{\text{\em\tiny U}}\:$}}
\newcommand{\unifnsp}{\text{$\vdash_{\text{\em\tiny U}}$}}
\newcommand{\unifsp}{\text{$\vdash_{\text{\em\tiny U}}$} }

\newcommand{\munif}{\text{$\vdash_{\tiny \unifcalc}\:$}}
\newcommand{\munifnsp}{\text{$\vdash_{\tiny \unifcalc}$}}
\newcommand{\munifsp}{\text{$\vdash_{\tiny \unifcalc}$} }

\newcommand{\lmndz}{\text{\lm--\cndz}}
\newcommand{\lmndzsp}{\text{\lm--\cndz} }

\newcommand{\lmhilz}{\text{\lm--$H_0$}}
\newcommand{\lmhilzsp}{\text{\lm--$H_0$} }

\newcommand{\lmndo}{\text{\lm--\cndo}}
\newcommand{\lmndosp}{\text{\lm--\cndo} }

\newcommand{\lmdmo}{\text{$\tem{dm}_1$}}
\newcommand{\lmdmosp}{\text{$\tem{dm}_1$} }
\newcommand{\lmdmop}{\text{${\tem{dm}_1}'$}}
\newcommand{\lmdmopsp}{\text{${\tem{dm}_1}'$} }

\newcommand{\lmdmt}{\text{$\tem{dm}_2$}}
\newcommand{\lmdmtsp}{\text{$\tem{dm}_2$} }
\newcommand{\lmdmtp}{\text{${\tem{dm}_2}'$}}
\newcommand{\lmdmtpsp}{\text{${\tem{dm}_2}'$} }

\newcommand{\prem}{\tem{premise}}
\newcommand{\premsp}{\temv{premise} }

\newcommand{\pold}{\text{$P_{\tem{old}}$}}
\newcommand{\poldsp}{\text{$P_{\tem{old}}$} }

\newcommand{\pnew}{\text{$P_{\tem{new}}$}}
\newcommand{\pnewsp}{\text{$P_{\tem{new}}$} }


\newcommand{\cterm}[1]{\text{$#1$}}
\newcommand{\cform}[1]{\text{$#1$}}
\newcommand{\cquant}[1]{\text{$#1$}}


\newcommand{\cmneg}{\text{$\boldsymbol{\neg}$}}
\newcommand{\cmnegsp}{\text{$\boldsymbol{\neg}$ }}
\newcommand{\cmnegnsp}{\text{$\boldsymbol{\neg}$}}
\newcommand{\cmand}{\text{$\:\boldsymbol{\wedge} \:$}}
\newcommand{\cmandsp}{\text{$\boldsymbol{\wedge}$ }}
\newcommand{\cmor}{\text{$\:\boldsymbol{\vee} \:$}}
\newcommand{\cmorsp}{\text{$\boldsymbol{\vee}$ }}
\newcommand{\cmif}{\text{$\:\boldsymbol{\Rightarrow}\,$}}
\newcommand{\cmifsp}{\text{$\:\boldsymbol{\Rightarrow}\,$ }}
\newcommand{\cmiff}{\text{$\:\boldsymbol{\Leftrightarrow} \,$}}
\newcommand{\cmiffsp}{\text{$\boldsymbol{\Leftrightarrow}$ }}
\newcommand{\cmandnsp}{\text{$\boldsymbol{\wedge}$}}
\newcommand{\cmornsp}{\text{$\boldsymbol{\vee}$}}
\newcommand{\cmifnsp}{\text{$\boldsymbol{\Rightarrow}$}}
\newcommand{\cmiffnsp}{\text{$\boldsymbol{\Leftrightarrow}$}}

\newcommand{\cforall}{\text{$\boldsymbol{\forall}$}}
\newcommand{\cforallsp}{\text{$\boldsymbol{\forall}$} }
\newcommand{\cexists}{\text{$\boldsymbol{\exists}$}}
\newcommand{\cexistsp}{\text{$\boldsymbol{\exists}$} }

\newcommand{\cforallLst}{\text{$\boldsymbol{\forall}^*$}}
\newcommand{\cforallLstsp}{\text{$\boldsymbol{\forall}^*$} }
\newcommand{\cexistsLst}{\text{$\boldsymbol{\exists}^*$}}
\newcommand{\cexistsLstsp}{\text{$\boldsymbol{\exists}^*$} }

\newcommand{\freshv}{\lmcon{fresh\text{-}var}}
\newcommand{\freshvsp}{\lmcon{fresh\text{-}var}\msp}

\newcommand{\cuspec}{\lmcon{uspec}}
\newcommand{\cuspecsp}{\lmcon{uspec} }

\newcommand{\cegen}{\lmcon{egen}}
\newcommand{\cegensp}{\lmcon{egen} }

\newcommand{\cqsaxiom}{\lmcon{qs\text{-}axiom}}
\newcommand{\cqsaxiomsp}{\lmcon{qs\text{-}axiom} }

\newcommand{\necrule}{\mbf{nec:}}
\newcommand{\necrulesp}{\mbf{nec:} }
\newcommand{\rulek}{\mbf{K}}
\newcommand{\ruleksp}{\mbf{K} }

\newcommand{\rulet}{\mbf{T}}
\newcommand{\ruletsp}{\mbf{T} }

\newcommand{\ruled}{\mbf{D}}
\newcommand{\ruledsp}{\mbf{D} }

\newcommand{\rulesfour}{\text{$\mbf{S}_4$}}
\newcommand{\rulesfoursp}{\text{$\mbf{S}_4$} }
\newcommand{\rulesfive}{\text{$\mbf{S}_5$}}
\newcommand{\rulesfivesp}{\text{$\mbf{S}_5$} }

\newcommand{\systemsfour}{\text{$\mc{S}_4$}}
\newcommand{\systemsfoursp}{\text{$\mc{S}_4$} }
\newcommand{\systemsfive}{\text{$\mc{S}_5$}}
\newcommand{\systemsfivesp}{\text{$\mc{S}_5$} }
\newcommand{\systemk}{\mc{K}}
\newcommand{\systemksp}{\mc{K} }
\newcommand{\systemd}{\mc{D}}
\newcommand{\systemdsp}{\mc{D} }
\newcommand{\systemt}{\mc{T}}
\newcommand{\systemtsp}{\mc{T} }
\newcommand{\necop}{\text{$\Box$}}
\newcommand{\necopsp}{\text{$\Box$} }
\newcommand{\posop}{\text{$\Diamond$}}
\newcommand{\posopsp}{\text{$\Diamond$} }
\newcommand{\nec}[1]{\text{$\necop #1$}}
\newcommand{\pos}[1]{\text{$\posop #1$}}

\newcommand{\kframe}[1]{\mc{#1}}
\newcommand{\kframesp}[1]{\mc{#1} }

\newcommand{\kprovesls}{\text{$\:\vdash_{\text{\scriptsize \systemk}}\:$}}
\newcommand{\kproves}{\text{$\vdash_{\text{\scriptsize \systemk}}$}}
\newcommand{\kprovesp}{\text{$\vdash_{\text{\scriptsize \systemk}}$} }
\newcommand{\nkprovesls}{\text{$\:\not{\hspace*{-0.007in}\vdash_{\scriptsize \systemk}}\:$}}

\newcommand{\posintro}{\text{\posop\mbf{-intro}}}
\newcommand{\posintrosp}{\text{\posop\mbf{-intro}} }
\newcommand{\poselim}{\text{\posop\mbf{-elim}}}
\newcommand{\poselimsp}{\text{\posop\mbf{-elim}} }

\newcommand{\propappfun}{\mc{F}}
\newcommand{\propappfunsp}{\mc{F} }

\newcommand{\lmirel}{\text{$\equiv$}}
\newcommand{\lmirells}{\text{$\:\equiv\:$}}
\newcommand{\lmirelsp}{\text{$\equiv$} }

\newcommand{\canonassign}{\text{$\rho_{\mc{S}}$}}
\newcommand{\canonassignsp}{\text{$\rho_{\mc{S}}$} }

\newcommand{\extendsls}{\text{$\:\succeq\:$}}
\newcommand{\properlyextendsls}{\text{$\:\succ\:$}}
\newcommand{\modalextends}{\text{$\succeq$}}
\newcommand{\modalextendsp}{\text{$\succeq$} }
\newcommand{\modalproperlyextends}{\text{$\succ$}}
\newcommand{\modalproperlyextendsp}{\text{$\succ$} }

\newcommand{\nextendsls}{\text{$\:\not\succeq\:$}}
\newcommand{\nproperlyextendsls}{\text{$\:\not\succ\:$}}

\newcommand{\ttm}[1]{\(#1\)}
\newcommand{\ttlam}{\ttm{\lambda}}
\newcommand{\ttsub}[2]{\(#1\sb#2\)}
\newcommand{\esb}[1]{\ttsub{e}{#1}}

\newcommand{\ttp}{\ttm{\Pi}}
\newcommand{\piof}[1]{\ttm{\Pi}\ttm{#1}}

\newcommand{\nequal}{\text{$\approx$}}
\newcommand{\nequalsp}{\text{$\approx$} }
\newcommand{\nequals}{\text{$\:\approx\:$}}
\newcommand{\notnequal}{\text{$\not\approx$}}
\newcommand{\notnequals}{\text{$\:\not\approx\,$}}

\newcommand{\neclog}{\mc{TL}}
\newcommand{\neclogsp}{\mc{TL} }

\newcommand{\utran}[1]{\text{$\mathfrak{U}\,(#1)$}}

\newcommand{\utransp}{\text{$\mathfrak{U}$} }

\newcommand{\utrannsp}{\text{$\mathfrak{U}$}}


\newcommand{\ptran}[1]{\text{$\mathfrak{P}(#1)$}}

\newcommand{\ptransp}{\text{$\mathfrak{P}$} }

\newcommand{\ptrannsp}{\text{$\mathfrak{P}$}}


\newcommand{\rctran}[1]{\text{$\mathfrak{C}(#1)$}}

\newcommand{\rctransp}{\text{$\mathfrak{C}$} }

\newcommand{\rctrannsp}{\text{$\mathfrak{C}$}}


\newcommand{\rasstran}[1]{\text{$\mathfrak{RL}(#1)$}}

\newcommand{\rasstransp}{\text{$\mathfrak{RL}$} }

\newcommand{\rasstrannsp}{\text{$\mathfrak{RL}$}}


\newcommand{\htran}[1]{\text{$\temls{Hoist}(#1)$}}

\newcommand{\htransp}{\text{$\temls{Hoist}$} }

\newcommand{\htrannsp}{\text{$\temls{Hoist}$}}


\newcommand{\mstran}[1]{\text{$\mathfrak{MS}(#1)$}}

\newcommand{\mstransp}{\text{$\mathfrak{MS}$} }

\newcommand{\mstrannsp}{\text{$\mathfrak{MS}$}}


\newcommand{\atran}[1]{\text{$\mathfrak{A}_1(#1)$}}
\newcommand{\atransp}{\text{$\mathfrak{A}_1$} }
\newcommand{\atrannsp}{\text{$\mathfrak{A}_1$}}

\newcommand{\aatran}[1]{\text{$\mathfrak{A}_2(#1)$}}
\newcommand{\aatransp}{\text{$\mathfrak{A}_2$} }
\newcommand{\aatrannsp}{\text{$\mathfrak{A}_2$}}

\newcommand{\aaatran}[1]{\text{$\mathfrak{A}_3$}}
\newcommand{\aaatransp}{\text{$\mathfrak{A}_3$} }
\newcommand{\aaatrannsp}{\text{$\mathfrak{A}_3$}}

\newcommand{\normalize}[1]{\temls{normalize}(#1)}

\newcommand{\normalizesp}{\text{\em normalize\/} }

\newcommand{\normalizensp}{\text{\em normalize}}

\newcommand{\restructure}[1]{\temls{restructure}(#1)}

\newcommand{\restructuresp}{\text{\em restructure\/} }

\newcommand{\restructurensp}{\text{\em restructure}}

\newcommand{\contract}[1]{\temls{contract}(#1)}

\newcommand{\contractsp}{\text{\em contract\/} }

\newcommand{\contractnsp}{\mbox{\em contract}}

\newcommand{\marr}[1]{\text{$\underrightarrow{\vspace*{0.048in}\:#1\:}$}}

\newcommand{\etour}{\temls{detour?}}
\newcommand{\etoursp}{\temv{detour?} }
\newcommand{\etournsp}{\tem{detour?}}

\newcommand{\rr}[1]{\text{$\tem{RR}(#1)$}}

\newcommand{\rl}[1]{\text{$\tem{RL}(#1)$}}

\newcommand{\even}[1]{\mbox{$\temv{Even}(#1)$}}
\newcommand{\odd}[1]{\mbox{$\temv{Odd}(#1)$}}
\newcommand{\parity}{\mbox{\mc{P}\hspace*{-0.03in}\mc{AR}}}
\newcommand{\paritysp}{\mbox{\mc{P}\hspace*{-0.03in}\mc{AR}} }
\newcommand{\meanfun}{\mc{M}}
\newcommand{\meanfunsp}{\mc{M} }
\newcommand{\mean}[1]{\mbox{$\denot{\mc{M}}{#1}$}}
\newcommand{\meantwo}[2]{\mbox{$\denot{\mc{M}}{#1}\:#2$}}
\newcommand{\deddom}{\mbf{Ded}}
\newcommand{\deddomsp}{\mbf{Ded} }
\newcommand{\abspace}{\mbf{AB}}
\newcommand{\abspacesp}{\mbf{AB} }
\newcommand{\propdom}{\mbf{Prop}}
\newcommand{\propdomsp}{\mbf{Prop }}
\newcommand{\evrule}{\mbf{even-next}}
\newcommand{\evrulesp}{\mbf{even-next} }
\newcommand{\zeroevrule}{\mbf{zero-even}}
\newcommand{\zeroevrulesp}{\mbf{zero-even} }
\newcommand{\zeroevrulename}{\tem{ZeroEven}}
\newcommand{\zeroevrulenamesp}{\tem{ZeroEven} }
\newcommand{\odrule}{\mbf{odd-next}}
\newcommand{\odrulesp}{\mbf{odd-next} }
\newcommand{\evrulename}{\temv{EvenNext}}
\newcommand{\evrulenamesp}{\temv{EvenNext} }
\newcommand{\odrulename}{\temv{OddNext}}
\newcommand{\odrulenamesp}{\temv{OddNext} }
\newcommand{\comprulename}{\temv{Comp}}
\newcommand{\comprulenamesp}{\temv{Comp} }
\newcommand{\claimrulename}{\temv{Claim}}
\newcommand{\claimrulenamesp}{\temv{Claim} }

\newcommand{\cndzprovesls}{\mbox{$\:\vdash_{\mbox{\scriptsize \cndz}}\:$}}
\newcommand{\cndzproves}{\mbox{$\vdash_{\mbox{\scriptsize \cndz}}$}}
\newcommand{\cndzprovesp}{\mbox{$\vdash_{\mbox{\scriptsize \cndz}}$} }
\newcommand{\ncndzprovesls}
{\mbox{$\:\not{\hspace*{-0.007in}\vdash_{\scriptsize \cndz}}\:$}}

\newcommand{\truerulename}{\temv{True}}
\newcommand{\truerulenamesp}{\mbox{\em True\/} }
\newcommand{\falserulename}{\temv{False}}
\newcommand{\falserulenamesp}{\mbox{\em False\/} }

\newcommand{\assumerulename}{\temv{Assume}}
\newcommand{\assumerulenamesp}{\temv{Assume} }

\newcommand{\pab}{\mbox{$\phi$}}
\newcommand{\pabsp}{\mbox{$\phi$} }

\newcommand{\buildcert}{\smtt{generate-certificate}}
\newcommand{\buildcertsp}{\smtt{generate-certificate} }

\newcommand{\buildcerttext}{\text{generate-certificate}}
\newcommand{\buildcerttextsp}{\text{generate-certificate} }

\newcommand{\opair}[2]{\mbox{$(#1,#2)$}}
\newcommand{\otriple}[3]{\mbox{$(#1,#2,#3)$}}
\newcommand{\oquadruple}[4]{\mbox{$(#1,#2,#3,#4)$}}
\newcommand{\oquintuple}[5]{\mbox{$(#1,#2,#3,#4,#5)$}}
\newcommand{\oseptuple}[6]{\mbox{$(#1,#2,#3,#4,#5,#6)$}}

\newcommand{\wrt}{\mbox{w.r.t.} }

\newcommand{\gac}{\mbox{\mtt{getAllCompletions}}}
\newcommand{\gacsp}{\mbox{\mtt{getAllCompletions}} }


\section{Introduction}

The Bloomberg Professional Service, popularly known as the Terminal,
has been a leading source of financial data, analytics, and insights
for over 30 years. Customers use it to query a wide variety 
of structured, semi-structured, and unstructured sources, create alerts, plot
charts, draw maps, compute statistics, \etcsp
For most of its history,  queries to the terminal have been built 
via dedicated GUIs. For example, if users wanted to find bonds 
that satisfied certain criteria, they would first need to 
navigate to a bond-search function,
and then specify the conditions of interest by interacting with a variety 
of GUI widgets. Long-time power users of the
Terminal are typically comfortable with their usual workflows. However,
the large number of available functions and the complex GUI interactions
they require may present challenges to those who need to step outside
their usual workflows, and can impose a learning curve on newcomers~\cite{DBLP:conf/sigmod/JagadishCEJLNY07,DBLP:journals/debu/LiJ12}. 

To mitigate these challenges, we have undertaken work
aimed at allowing users to interact with the Terminal in natural language,
and specifically to formulate queries directly in natural language.
These range from simple factoid questions to structurally
complex queries. The following are representative examples from a number of different
domains:
\begin{itemize}
  \setlength\itemsep{0.01em}
  \item {\em What are the top 5 European auto companies with eps at least 3?}
  \item {\em What was Apple's market cap in the second quarter?}
  \item {\em Show me investment grade bonds in the emerging markets with yield $>$ 4\%}
  \item {\em News about brexit from the New York Times between September and now}
  \item {\em Chart a histogram of Netflix vs AT\&T cable subscribers over the last 5 years}
  \item {\em Tech CEOS in California who graduated from Harvard Business School}
\end{itemize}
Our QA (question answering \cite{DBLP:conf/cikm/BastH15,DBLP:journals/sigmod/LiJ16,DBLP:conf/acl/SavenkovA17}) systems
use semantic parsing to compute a formal representation of the meaning of such
a query. These representations are then translated into executable query languages 
(such as SQL or SPARQL). Those queries are finally 
executed against the back end and the results are presented to the user.

However, natural language interfaces present usability challenges of their own~\cite{Helander:1997:HHI:549940}.
In short, it is not clear to users what a QA system can and cannot do. 
The first part pertains to {\em discovery}, and specifically to discovering
what the system {\em can\/} do---what class of questions or commands 
it can understand. The second part pertains to
{\em expectation management}: We want to steer the users away from the (inevitable)
limitations of the QA system. Such limitations include lack of support for
specific kinds of functionality, incompleteness of the underlying data,
and limitations of semantic parsing technology. 
We use auto-completion as a tool that can help to tackle both the
discovery problem, by suggesting queries which we know to be fully parsable
and answerable;  and expectation management, whereby we  stop offering suggestions
as a signal indicating that we are not able to understand and/or answer
what is being typed. Figure~\ref{Fig:SampleNewsQueries} showcases some 
inputs and outputs from our auto-complete system for news (a domain
we will discuss further in the sequel). 

\begin{figure}[t!]
\scriptsize
\flushleft
\setlength\tabcolsep{0.7pt}
\begin{tabular}{ | l | }
  \hline		
  amaz \hspace{64px}\, {\color{gray}1} \; \\
  \hline
  amazon \\
  amazon web services \\
  amazon japan \\
  amazon go \\
  amazon hq2 \\
  \hline  
\end{tabular}$\rightarrow$\begin{tabular}{ | l | }
  \hline			
  amazon news fro \hspace{57px}\,\, {\color{gray}2} \;\\
  \hline
  amazon news from the nyt \\
  amazon news from business insider \\
  amazon news from last week \\
  amazon news from today \\
  amazon news from april \\
  \hline  
\end{tabular}

\vspace {1mm}

\begin{tabular}{ | l | }
  \hline			
  amazon news from t\hspace{5mm} {\color{gray}3}\\
  \hline
  amazon news from the nyt \\ 
  amazon news from today \\
  amazon news from this year \\
  \hline  
\end{tabular}$\rightarrow$\begin{tabular}{ | l | }
  \hline			
  amazon news from thr \hspace{16mm} {\color{gray}4}\\
  \hline
  amazon news from three days ago \\
  amazon news from three months ago \\
  amazon news from three years ago \\
  \hline  
\end{tabular}
\caption{Semantic auto-completion in action.}
\label{Fig:SampleNewsQueries}
\end{figure}

Building AC (auto-complete) systems for new QA systems introduces
a set of unique challenges, the main one being 
the cold-start problem. Since AC systems aim to address 
fundamental usability issues with QA systems, we aim to release QA and AC
systems in tandem. This means that we don't have the luxury of large query
logs that can be used to bootstrap the AC systems. However,
on the positive side, because we have access to the grammatic structure
encoded in the semantic parser, it is possible, with the aid of appropriate lexicons
and certain statistics, to generate large sets of queries synthetically,
which can then be used as if they were user queries. Synthetically
generated queries will never be as good as the genuine article, but when
carefully prepared they can be very helpful.

In this paper we report on our experience building AC systems for natural language
interfaces. Specifically, the paper makes the following contributions:
\begin{itemize}
  
\item we introduce the problem of auto-completion for QA systems that are
  based on semantic parsing, and identify a set of properties that systems
  tackling this problem should satisfy (Section~\ref{Sec:ProblemStatement});
        
  \item we outline our approach and a number of algorithms that we use to tackle this problem (Section~\ref{Sec:Approaches});
    
  \item we report experimental results on the effectiveness and efficiency of the AC systems  
    we have been building at Bloomberg (Section~\ref{Sec:ExpResults}).
    
\end{itemize}
The following section provides some brief background on
semantic parsing (Section~\ref{Sec:SemParsingBackground}), and the last section 
discusses related work (Section~\ref{Sec:RelatedWorkAndConclusions}).


\section{Background: Semantic Parsing}
\label{Sec:SemParsingBackground}


\begin{figure*}[ht]
\hspace{5mm}$q$ = \textit{``chinese non-tech bonds maturing in three years''}

\vspace{1mm}

\hspace{5mm}$\phi$ = \smtt{(COUNTRY_OF_RISK = CHINA) AND NOT(SECTOR = SEC_TECH) AND MATURITY_DATE = RELATIVE_TIME(3,YEAR,NOW)}

\vspace{1mm}

\noindent\hspace{5mm}$D(q,\phi)$ =

\small
\centering
\vspace*{3mm}
\shadowbox{\begin{minipage}{56em}
\begin{forest}
[$\phi$
	[$\phi_1\;$ \\ \smtt{(COUNTRY_OF_RISK = CHINA)}
		[\smtt{CHINA}, tier=preterminal
		 [\textit{chinese}, tier=terminal]
		]
	]
	[$\phi_2\;$ \\ \smtt{NOT(SECTOR = SEC_TECH)}
		[
			[\smtt{NOT}, tier=preterminal
				[\textit{non}, tier=terminal]
			]
			[\smtt{\msp \msp \rsp \rsp (SECTOR =$\;$}\\\smtt{SEC_TECH)}
				[\smtt{SEC_TECH}, tier=preterminal
					[\textit{tech}, tier=terminal]
				]
			]
		]
	]
	[$\phi_3\;$ \\ \smtt{(MATURITY_DATE = (RELATIVE_TIME(3,YEAR,NOW))}
		[\smtt{MATURITY_DATE}, tier=preterminal
			[\textit{maturing}, tier=terminal]
			[\textit{in}, tier=terminal]
		]
	    [\smtt{3}, tier=preterminal
				[\textit{three}, tier=terminal]
		]
		[\smtt{YEAR}, tier=preterminal
				[\textit{years}, tier=terminal]
		]
	]
]
\end{forest}
\end{minipage}}
\caption{A bonds query with its interpretation and corresponding derivation.}
\label{Fig:SampleDerivation}
\end{figure*}
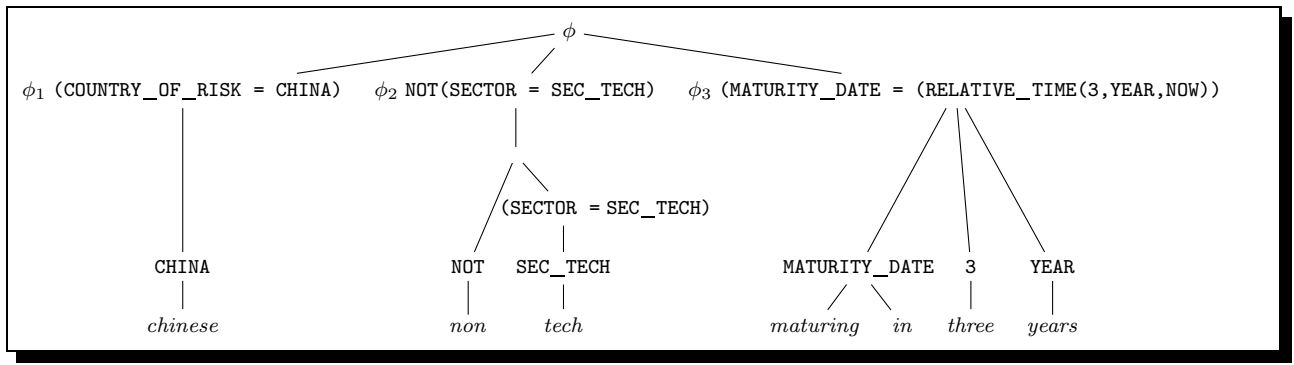

Semantic parsers map natural language utterances into logical forms
that capture their meaning~\cite{DBLP:journals/corr/abs-1812-00978}. This is done in two conceptual stages.
The first is a \emph{parsing analysis}, whereby a sentence is mapped
to all interpretations that can be derived from it, 
reflecting the lexical and syntactic ambiguity of natural language. The second 
is a \emph{ranking} stage aimed at ordering these interpretations in accordance with their plausibility. 
The top interpretation is then executed by the back end in order to return results,
plot a chart, set up an alert, \etcsp 

We provide QA systems based on semantic parsing for multiple domains, each of which might have its own domain-specific executable query language
in the back end, reflecting differences in the underlying data models and supported operations. It is not feasible
to tailor each QA system to each desired target query language. Therefore, we use a generic 
\emph{intermediate representation} language (IR) based on a fragment of first-order logic.
Our semantic parsers map natural language to this IR, and IR formulas are then
readily translated to whatever executable query language is used by the domain's back end.

An IR formula $\phi$
is typically either an atomic formula $\alpha$ (or {\em atom\/} for short);  
or else a complex sentential combination of formulas, namely, negations, conjunctions,
or disjunctions. Atoms are usually equalities between variables (or {\em fields,} also
known as {\em attributes}) and {\em values\/}, where a variable has a primitive type,
usually numeric (integer or real); or an {\em enumeration\/} (\temv{enum} for short), in which case the value must be
one of the finitely many values of that variable; or string; or boolean. Examples of
numeric fields are stock price and salary; examples of enumeration fields include credit ratings and 
country of domicile; and examples of boolean fields are actively traded (for tickers), convertible (for bonds),
privately held (for companies), etc. An atom might also be an inequality such as $f \leq v$ or $f > v$, 
when $f$ is a numeric field and $v$ is a numeric value (possibly with units or other
additional information attached). In the general case, an atom can be any $n$-ary 
relation between values of certain types, for $n > 0$. 
Here we will be mostly concerned with 
formulas that are conjunctions of one or more atoms:
$\phi \equiv \alpha_1 \mand \cdots \wedge \alpha_n$ where $n > 0$.
Informally, a derivation 
$D(q, \phi)$ is a tree whose terminal nodes are the tokens in $q$ and whose non-terminal nodes correspond to
logical and non-logical symbols and formulas. A derivation is shown in  Figure~\ref{Fig:SampleDerivation}.

We do not have space here to say much about our semantic parsing technology, but it is important to note
that our discussion in this paper is agnostic on that point. 
The semantic parsers could be based on CCGs and machine learning, or on PCFGs and first-order or higher-order
logic (with or without machine learning), or on parser combinators, or even on a purely
deep learning pipeline. The only requirement is that there should be some notion of a structured
(tree-based) meaning representation.


\section{Problem Statement}
\label{Sec:ProblemStatement}
We now outline a set of properties that should be satisfied by AC systems
designed to improve the usability of semantics-based QA technology. 
The problem we address in this paper is building AC systems that satisfy these properties. 
As the QA and corresponding AC systems should ideally be released together, a major challenge
that we deal with is satisfying these properties despite the scarcity of data in the form 
of query logs resulting from cold starts.

Two minimal requirements that characterize the AC problem in our setting 
are {\em soundness\/} and {\em completeness}. 
Soundness itself is split into two properties, syntactic and semantic soundness. 
An AC system is syntactically sound provided that every completion $q = t_1 \cdots t_m$ that
it returns for a given partial query $p = s_1 \cdots s_n$ is a syntactic
{\em extension\/} of $p$, meaning that the tokens of $q$ can be partitioned into two sets $T_1$
and $T_2$, where every $t_i \in T_1$ is a suffix of a unique token $s_j$ in $p$, and 
every $s_j$ in $p$ is a prefix of a unique $t_i \in T_1$. It is
semantically sound if $q$ is semantically parsable and answerable.
Both properties are conditional statements and thus would be easy to attain 
if the system never provided any completions. We also need completeness:
The system should provide at least some completions whenever 
$p$ is in fact extensible to some semantically parsable and answerable query. 

But we need a number of additional properties above and beyond soundness and
completeness: 
\begin{enumerate}
\item The completions should be {\em predictive\/} of user intent; 
  in particular, the user's intended query should be as high up on the list of completions as possible.
\item The completion list should be {\em diverse}: It should contain entries
  of different types.
  In the case of a QA system for news, for example, if the partial query is the letter $i$, 
  the results should not be limited to companies
  whose names  start with an {\em i}, such as IBM; it should include people (such as Icahn),
  sectors (such as insurance), regions (Ireland), and so on. Diversity is very important for discoverability.
\item The completions should be {\em propositional}, meaning that they should have full
  sentential semantics: The semantic parser must fully map the completion to a formula
  in the underlying logic, which could be a sentential atom or a more complex formula.
  For example, if the partial query is 
  {\em investment grade bonds i}, then {\em investment grade bonds in the emerging markets\/}
  is an acceptable propositional completion, but {\em investment grade bonds in the\/} is not.
  All completions in Figure~\ref{Fig:SampleNewsQueries} are propositional.
\item The completions should be as {\em grammatical\/} as possible, modulo what the user has already typed. 
  The QA system should be able to understand telegraphically formulated queries \cite{DBLP:conf/emnlp/JoshiSC14},
  but nevertheless we should strive to offer completions that are as linguistically well-formed as possible.
  There is tension between this requirement and completeness, which is why we formulate this as a soft constraint.

\end{enumerate}

The above properties are our main focus, but there are other desiderata as well,
such as personalization (the history and profile
of a user should affect the completions they receive) and popularity (popular
queries posed by other users should be more likely to appear as completions) \cite{DBLP:conf/sigir/Shokouhi13}.


\section{Approach}
\label{Sec:Approaches}
We now outline the high-level approach we take to solve the auto-completion problem
introduced and motivated above. The approach relies on a number of different 
{\em completion algorithms}, each of which takes a \mtt{prefix} string provided
by the user ($p$ in the notation of Section \ref{Sec:ProblemStatement}), potentially
along with additional domain-dependent configuration parameters, and returns
an ordered sequence of \mtt{Completion} objects, each of which
contains at least four pieces of information:

\squishlist
\item \mtt{completion}: the completion string to be shown to the user ($q$ in Section \ref{Sec:ProblemStatement}).
  
  \item \mtt{interpretation}: the interpretation (semantics AST) of the \mtt{completion}. 
        The interpretation is used for deduplicating completions. In some instances, a simplified form of the
        interpretation may be shown to the user to help disambiguate the meaning of the completion, or 
        introduce the user to a domain's vocabulary.
  \item \mtt{type}: one of a finite number of identifiers designating the different types of completions, 
         used for maximizing diversity. 
  \item \mtt{grade}: a qualitative score comparable across different completion algorithms and used for weaving 
        and ranking the final set of completions.
\squishend

When an AC system receives an input prefix, it passes it to a \emph{top-level coordinating algorithm} that
runs a number of available completion algorithms in parallel (we describe the main algorithms in the following sections).
The coordinating algorithm will wait until all algorithms return their completions. 
Each individual algorithm is expected to return a diversified ranked list of completions. 
The coordinating algorithm then takes these lists and weaves them in a way that again ensures diversity
and respects the grades. The top-level algorithm will also ensure that semantic and lexical duplicates are eliminated,
though the individual (lower-level) algorithms may also perform their own deduplication. Note that the 
availability of semantics allows us to detect duplicate completions in a much stronger sense than would be allowed
by simple morphology. For instance, \temv{ibm} and \temv{big blue} will be conflated assuming that their
semantic representation is identical (the ticker \mtt{IBM}). 
Coordinating algorithms can be customized on a 
per-domain basis. For example, in some domains it might make sense 
to run some of the algorithms only if all other algorithms fail to return results.
In other domains, completions with low grades may be eliminated if there are completions with high or medium
grades. 
We now proceed to describe the main completion algorithms that are typically used in most domains.


\subsection{Most Popular Completion (\mbf{mpc})}
\label{Sec:TopAlgo}
A natural starting point for auto-completion is utilizing available query logs
(following a number of careful steps intended to safeguard user privacy).
As discussed, in our setting  we need to deploy a QA and a corresponding AC system in tandem,
which means we don't have any user query
logs initially. However, we can often use queries collected internally or from annotators
to provide a core initial set of queries for log-based AC, and we may also generate queries synthetically.

For fast matching against a user's partial query $q$, this algorithm uses standard \mbf{mpc} (``most popular completion'')
implementation techniques, albeit augmented to address aspects of the problem that are peculiar to our setting,
as discussed in Section~\ref{Sec:ProblemStatement}. 
The query log is stored in a trie $T_L$ where each query is paired with its frequency. 
We are typically interested in the top-$k$ matches with respect to the log frequency, where 
$k$ is typically around 50. These $k$ queries are then re-ranked using a domain-specific re-ranker.

As discussed earlier, each completion needs to be annotated with its interpretation,  type identifier (for diversification),
and grade. All three pieces of information are pre-computed offline when query logs are ingested. Pre-computation 
allows us to keep the amount of computation done online to a minimum. 
Diversification information is computed offline for all atomic constraints in the interpretation
of a query. Online, at completion time, the constraint whose type is used as the diversification
identifier of the completion is determined dynamically based on the user's prefix with respect to the full completion.
Given a prefix $p$ and a completion $q_p$, where $p$ is a prefix of $q_p$ and $\phi$ is the interpretation of $p_q$ with
atomic constraints $\{\phi_1, \phi_2,...,\phi_n\}$, the diversification class is a function of $\phi_p$, the constraint with a 
path to the last token in $p$ in the derivation tree of $q_p$. This simply means that diversification is performed 
based on the constraint that is currently being typed by the user. For example, the user prefix 
$p=$\textit{``chinese non-te''} can be completed to $q_p=$ \textit{``chinese non-tech bonds maturing in three years''},
whereby the last token in $p$ is \textit{``te''}. If we look at the derivation of the interpretation of $q_p$ in 
Figure~\ref{Fig:SampleDerivation}, the completion of that token in the prefix corresponds to the 
\smtt{SECTOR} constraint, which means that  \smtt{SECTOR} is used as the diversification type for this completion.

In some cases, queries from the logs cannot be suggested in their raw form, requiring  some type of reformulation.
For example, a query like \textit{``list all chinese tech bonds that will mature between April 1, 2018 and May 30, 2020''}
is meaningful as a completion up until April 1, 2018, but not after. The issue stems from the combination of tense and
explicit time expressions. We will describe how we deal with this issue as an example of the general problem  of 
\emph{query log reformulation}. We deal with issues caused by explicit time expressions in the query log
by a time-shifting reformulation. In this setting, we have the logged query $q$ observed at time $t_q$ that contains 
explicit references to times $\{t_{e_1}, t_{e_2},...,t_{e_n}\}$ and the current time at which the 
time-shift reformulation is happening is $t_{\mathrm{now}}$. The shifted times are obtained as follows:
$t'_{e_i} = t_{e_i} + t_{\mathrm{now}} - t_q$. 
The result of performing this shift with $t_{\mathrm{now}} = \text{January 1, 2020}$  
and $t_q = \text{January 1, 2018}$ is 
\textit{``list all chinese tech bonds that will mature between April 1, 2021 and May 30, 2022''}. The shift
is based on the observation that users are typically interested in information related to a relative time in the 
future or past, but expressed using absolute times (rather than relative ones like \textit{``yesterday''} or \textit{``in two months''}).

\subsection{Atomic Completions (\mbf{atomic})}
\label{Sec:atomic}

As described above, log-based completion suffers from a major shortcoming that
results in underutilization of query logs. In particular, a completion provided 
by this algorithm must be {\em a log query that contains all tokens in the user's partial query},
\iensp, a log query that extends whatever the user has typed.
This is an exceedingly strong condition. As a very simple example, suppose that
our log contains only the following two queries: 
\begin{verbatim}
ibm bonds maturing in 2020
bullet bonds with yield > 2 pct 
\end{verbatim}  
And suppose now that the user types the partial query
\begin{equation}
  \temv{bullet bonds mat}
  \label{Eq:AtomiCompEx1}
  \end{equation}
The log-based algorithm that we have described is incapable of offering any completions
for input~(\ref{Eq:AtomiCompEx1}), 
because {\em there is no single query in the log that extends this input}. The first
query in our log does not contain {\em bullet}, while the second one does not
contain any tokens that extend {\em mat}.

Note that this inability persists even
if we loosen up the notion of matching used by the log-completion algorithm.
For instance, we might not insist on left-to-right matching, so that an input like {\em in 2020\/} might
still be completed to the first query, {\em ibm bonds maturing in 2020}. But even
with such restrictions lifted, the algorithm would be unable to complete~(\ref{Eq:AtomiCompEx1}),
because the gist of the limitation is that no {\em one single query\/} in the log
matches~(\ref{Eq:AtomiCompEx1}). The fundamental units returned by this algorithm
are {\em entire queries\/} in the log, which is not just unduly limiting as just indicated,
but is also often inappropriate. For example, if the user types {\em ib\/}, the log-based 
algorithm will return {\em ibm bonds maturing in 2020\/} as a completion, which is rather 
unnecessarily long and specific. Arguably, a more appropriate completion
might simply be {\em ibm bonds}. This is a more general completion (carries less
information than {\em ibm bonds maturing in 2020\/}) while still satisfying the
propositionality requirement. 

The {\em atomic completion algorithm\/} described in this section addresses these issues not by
completing to entire queries in the log, but rather to {\em atoms\/} found in the log,
or more precisely, to atom surface forms.\footnote{To simplify presentation, we will continue
to use the term ``atom'' to denote either a {\em logical atom}, \iensp, a piece of semantics
represented as an AST, 
or a phrase (a sequence of tokens) that has a logical atom as its semantics. The context will
always disambiguate the use.}
Typically, each query in the log contains multiple
atoms, so this might not only give
us a larger pool of potential results to return as completions, but the results themselves are
smaller, more general, and most importantly, they can be more flexibly
stitched on at the end of user inputs. In our simple running example, there are four
atoms total:
\begin{verbatim}
ibm bonds
maturing in 2020
bullet bonds
with yield > 2 pct 
\end{verbatim}  
The top two derive from the first query in the logs, and the bottom two from the second query. 
These four atoms now become our major completion candidates. The atoms we extract from a query
log are then organized into a trie $T_A$, the {\em atom trie}.
(We will have more to say on how atoms are extracted from query logs shortly.) 

Then, online, when we are given a partial query 
$p = w_1 \cdots w_k$ to complete, where $k \geq 1$ and $w_k$ is any sequence of characters that is a prefix of some token
in the vocabulary,\footnote{We ignore spelling correction in this paper.}
the atomic algorithm proceeds as follows: First, we use the domain's semantic parser to parse
as much of $p$ as possible. This means that our semantic parser must tolerate disfluencies and
noise, at least at the tail end of the input. This {\em decomposition analysis\/} splits the
input $p$ into two parts:
\ben
\item an {\em initial segment\/} $i_p = w_1 \cdots w_m$ that is understood by the semantic parser and results in
  some semantics $\phi_{i_p}$, where $m$ might be equal to $k$; and 
\item {\em the remainder\/} of the input, $r_p = w_{m+1} \cdots w_k$, which constitutes an {\em unrecognized segment}.
  \een
  Assuming that the remainder is non-empty, we match it against the atom trie $T_A$, and this returns
  a list of atoms $L = [A_1,\ldots,A_n]$ as potential completions. We then assign a score to each atom
  $A_j$, relative to the initial segment $i_p$. This score can be understood as a numeric measure of 
  the goodness of the fit between $A_j$ and $i_p$, \iensp, the degree to which $A_j$ is an appropriate
  completion of the unrecognized segment {\em given\/} or {\em conditioned upon\/} the existence of $i_p$
  to the left of $r_p$. 
  This is obtained by a scoring function $S$ that takes $i_p$, $r_p$, $A_j$, and an {\em atom model\/} $M$
  as inputs (to be described below). We then
  do a selection sort of $L$, based on $S(i_p,r_p,A_j,M)$, and for the desired number $d$ of completions
  (typically 5 or 10). For each of those top $d$ atoms, its untokenized form is appended to the end of
  $i_p$, and the result becomes the corresponding final atomic completion. The semantics of that final
  completion are typically obtained by conjoining $\phi_{i_p}$ with the (pre-cached) semantics of the
  atom $A_j$.\footnote{In rare cases, if the first word of $r_p$ is a logical connective such as {\em or},
    we might produce a disjunction instead of a conjunction.}

  As a quick illustration, suppose again that the input $q$ is {\em bullet bonds mat}. During the decomposition
  analysis, the semantic parser will recognize \tem{bullet bonds} as the maximal initial segment $i_p$ that
  is parsable, with atomic semantics
  \begin{equation}
    \mtt{MATURITY\_TYPE = BULLET}
    \label{Eq:AtCompSem1}
  \end{equation}
  and will identify the segment \tem{mat} as
  unrecognized. That segment will then be matched against the atom trie $T_A$ and will return the singleton list
  $[\temv{maturing in 2020}]$ as the only candidate completion (recall that  $T_A$ has only four atoms in the running
  example). So in this trivial example there is no ranking to be done, and by concatenating $i_p$ with this atom
  we obtain the one and only atomic completion: $$\temv{bullet bonds maturing in 2020}.$$ Its semantics are given
  by the conjunction of~(\ref{Eq:AtCompSem1}) and the semantics of the atom itself, namely
    \begin{equation}
    \mtt{MATURITY\_DATE = ExactDate(-1,-1,2020)}
    \label{Eq:AtCompSem2}
  \end{equation}
    (A date expression of the form \mtt{ExactDate(\mbox{\rm $d$},\mbox{\rm $m$},\mbox{\rm $y$})}, for integers $d$, $m$, and $y$, denotes the
    date corresponding to the respective coordinates, with a $-1$ indicating that no value was specified for
    the corresponding dimension. An expression of the form \mtt{Relative_Time($n$,$u$,$t$)}, for a temporal unit $u$,
    a numeric value $n$, and an anchor time expression $t$, indicates the time obtained by adding $n$ units $u$ to
    the time denoted by $t$.)

    How well this algorithm works hinges on the quality of the similarity metric $S$. 
  To describe how $S(i_p,r_p,A_j,M)$ is computed, we  first need to discuss the contents and offline generation
  of the atom model $M$. This model is essentially a map, computed offline and loaded upon initialization,
  from each atom $A_j$ to a record of
  information about $A_j$, such as its count (the number of times it occurs in the corpus),
  its tokenized and untokenized representation, its semantics (encoded as an AST), and most
  importantly, its {\em context vector space}. The context vector space of an atom $A_j$,
  denoted by $C_{A_j}$, is defined as a lexical vector space (sparse map from vocabulary words to integer counts),  
  obtained as follows:
  
  \bit
  \item Set $C_{A_j} = \eset$.
\item For every query $q$ in the log:
  \bit
\item For every occurrence of $A_j$ in $q$:
  \bit
  \item Let $w_1,\ldots,w_l$ be all and only
  the words in $q$ {\em to the left of that occurrence of $A_j$}. For each such $w_i$,
  set $$C_{A_j}[w_i] = C_{A_j}[w_i] + 1.$$
  \eit
  \eit
  \eit

  While this algorithm is parameterized over a single atom $A_j$, it is possible to
  build $C$ for all atoms at the same time with just one linear scan over all
  queries in the log. 

  To continue the running example, the model computed here would be of the following form
  (expressed in pseudo-JSON notation):
\begin{verbatim}
{"ibm bonds" :=  {
   semantics := "COMPANY_NAME = IBM",
   count := 1,
   context := {},
   ...
 },
 "maturing in 2020" := {
   semantics := "MATURITY_DATE = ExactDate(-1,-1,2020)",
   count := 1,
   context := {"ibm" := 1,
               "bonds" := 1},
   ...
 },
 "bullet bonds" := {
   semantics := "MATURITY_TYPE = BULLET",
   count := 1,
   context := {},
   ...
 },
 "with yield > 2 pct" := {
   semantics := "FLD_YLD > 2(PERCENT)"
   count := 1,
   context := {"bullet" := 1,
               "bonds"  := 1},
   ...
 }
}
\end{verbatim} 
That is, every atom $A_j$ contains its semantics, a {\em context\/} that encodes
the vector space $C_{A_j}$ above, and potentially a wealth of additional information
(such as linguistic features ranging from morphology to syntax and additional semantic
signals), represented above by ellipses. 

We can now outline the scoring function $S(i_p,r_p,A_j,M)$ as follows:
\bit

\item If $i_p$ and $A_j$ are incompatible, penalize $A_j$ (in proportion  
to the degree of incompatibility). This is typically determined by additional statistics that
are computed offline from the corpus and stored in the model $M$.

\item Otherwise, compute the similarity by looping through the words in $i_p$, using the context $C_{A_j}$
  as a grader:
  \bit
  \item Set $\temv{score} := 0.0$;
  \item For each $w \in i_p$:
    \bit
    \item $\temv{score} = \temv{score} + f(C_{A_j}[w])$
      \eit
    \item Return $h(\temv{score})$, where $h$ is a scaling function 
      (or the identity if no scaling is needed).
  \eit
  The function $f$ may be the identity function or some other layer of processing on top of the raw counts. 
  Accordingly,
   words in $i_p$ that have been seen before to the left of $A_j$ in the corpus are rewarded in proportion to
   how often they have been seen. Words in $i_p$ that have never been seen before to the left of $A_j$
   may be accordingly penalized.
\eit

To simplify presentation, we stated earlier that the remainder $r_p$ is matched against the atom trie $T_A$,
and that this operation returns a list of atoms $L = [A_1,\ldots,A_n]$ which are then ranked on the basis
of the scoring function. The real picture is only marginally more complicated, in order to ensure semantic diversity.
Specifically, the matching operation returns the list of trie matches $L$  partitioned into a set of {\em buckets}, where
all atoms in the same bucket have the same {\em type\/} of atomic semantics. That type is usually determined
by---and can be identified with---the field that occurs on the left-hand side of the operator \temv{op},
assuming that the atom is of the form $(\temv{field}\msp\temv{op}\msp\temv{value})$. For atoms
of different form, some other unique type identifier must be specified as the atom's ``type.''

These types play a dually useful role. First and foremost, they allow us to diversify the results by ensuring
that we don't get atomic completions of one type only (or two types only), say only completions of the form
$\mtt{COMPANY_NAME} = \cdots.$. 
This is particularly important for short inputs, because with hundreds of thousands of companies 
in total, there will be thousands of company names completing any one-letter prefix, and it might well be
that several of those will be popular. In general, we want to mix up the set of results to the greatest
possible extent while still ensuring that the provided completions are plausible and are reasonably predictive
of user intent. In our case this is accomplished by ranking each bucket separately and then weaving the
resulting ranked lists.

The second major use of atom types is in avoiding completions of a type that has already been encountered
in the initial segment $i_p$. For instance, consider the partial query $q = \temv{maturing in 2020 m}$. Here
$i_p$ is {\em maturing in 2020}, with semantics~(\ref{Eq:AtCompSem2}). As discussed, the type
of that atom can be identified with the field \mtt{MATURITY\_DATE}. The unrecognized
segment $r_p$ is the single letter \temv{m}, which will match very many atoms in $T_A$, quite a few of them
might be of the form $\temv{maturing in $\cdots$}$. We do not want to give a completion of the form
\begin{equation}
  \temv{maturing in 2020 maturing in 2023}
\label{Eq:BadAtom1}  
\end{equation}
or, even worse,
\begin{equation}
  \temv{maturing in 2020 maturing in 2020}.
\label{Eq:BadAtom2}
\end{equation}
Much more appropriate
completions might be $$\temv{maturing in 2020 issued by microsoft}$$
or $$\temv{maturing in 2020 mining sector}$$
and so on. While completions such as~(\ref{Eq:BadAtom1}) and (\ref{Eq:BadAtom2}) are naturally likely
to have lower scores due to the context analysis, it is much safer to weed them out of consideration
altogether by realizing that the type of the completion atom is identical to the rightmost type
of the initial segment (more precisely, the type of the rightmost atom in the
initial segment).\footnote{Depending on the domain, there are some fields and some constructions
  for which this type of juxtaposition is sensible and should not result in the elimination of
  the corresponding atom. For instance, a construction like ``french, german or italian bonds''
  necessitates the juxtaposition of three atoms of the same type, the first two of which are
  conjoined for prefixes prior to the ``or'' particle. These situations receive special treatment
  in our system on a configurable basis.}

  Note that the list $L = [A_1,\ldots,A_n]$ of candidate atoms that is obtained by matching
  $r_p$ against $T_A$ is already sorted, by a statically known measure such as the popularity
  of each $A_j$ (which may be defined simply as the number of occurrences of $A_j$ in the corpus).
  This is important for the following reason: In a domain with a large log of queries (which may
  be user queries or synthetically generated), there may be millions of atoms in the model, and a short
  unrecognized segment $q_r$ (\egnsp, only one character long) may return hundreds of thousands
  of atoms as candidates, or potentially even millions. The similarity function $S(i_p,r_p,A_j,M)$
  is fairly computationally expensive, so having to select the top 10
  or so atoms based on this function from a list $L$ of that length can be prohibitively expensive.
  A lot of work can be saved here by realizing that in such cases (very short prefixes that match 
  a very large number of candidate atoms) we are dealing with an embarrassment of riches: 
  If $L$ is already pre-sorted by sheer atom count, then we can simply take the top 100K or so atoms in
  the front of the list and drop the remainder. The most popular 100K atoms are guaranteed
  to give us the results we want for that kind of input: more than enough popular atoms 
  and with more than enough semantic diversity.
  Atoms in the tail of the list are much less likely
  as completions at that point; if needed, they will surface subsequently as the user types
  additional characters. 

  Occasionally, the initial split produced by the decomposition analysis is not optimal.
  In particular, the initial segment  $i_p$ may be overly long, and to get the right split
  we need to backtrack, by shifting one or more tokens from the tail end of $i_p$ to the
  front end of $r_p$. This might happen when $r_p$ is initially empty (and thus $i_p$ is identical
  to $q$), or it might happen when both $i_p$ and $r_p$ are nonempty. As an example of the
  first case, consider a partial query like \temv{ibm b}. Because $b$ is a credit rating and our semantic
  parser is by necessity flexible in order to understand queries that are telegraphically or
  elliptically expressed, this entire query is fully parsed as
  $$\mtt{COMPANY_NAME} = \mtt{IBM} \mand \mtt{BB\_COMPOSITE\_RATING} = \mtt{B},$$
  which would mean that there is no unrecognized segment to match against $T_A$ and therefore
  no completions offered by the atomic algorithm. A better decomposition analysis, however,
  is to treat the entire partial query as unrecognized, setting $i_p$ to the empty sequence.
  This would result in a wealth of atomic completions like \temv{ibm bullet}, \temv{ibm bonds}, \temv{ibm b or better},
  and so on. We have heuristics in place for determining when and how far to backtrack in such a case.
  In the second case, when both $i_p$ and $r_p$ are nonempty, a simple but effective heuristic for
  evaluating the goodness of the split is the number of results returned by the trie match. If we get a small  
  number of results from the initial split, but a much larger number when we backtrack by a certain number of tokens,
  then the latter split should be preferred. Note that backtracking cannot proceed on a token-by-token
  basis, because we need to maintain the invariant that $i_p$ is fully recognized and its semantics consist
  of a number of constraints. Therefore, on each backtracking attempt, we need to backtrack by
  the exact number of tokens that correspond to the rightmost constraint left in $i_p$ at that point. 
  Ideally, of course, we would consider all possible decompositions, obtain completions for each of them,
  and then merge in accordance with the scores. However, such an approach would be time-inefficient. Our results
  indicate that the present approach of picking one single decomposition to work with, but using informed heuristics
  in its selection, provides high-quality results and is efficient. 
  
  We close this section by pointing out that while the notion of an atom is typically fixed by the semantics of the domain at hand, 
  with sufficient imagination it is possible to extend that notion to include altogether different types of atoms, 
  thereby rendering the atomic completion algorithm applicable in novel ways. Indeed, it is possible
  to have multiple instances of the atomic algorithm running in parallel (see Section~\ref{Sec:Approaches}),  
  one that is based on the conventional notion
  of atom, as determined by the usual semantics of the domain at hand, and others based \label{Pg:NewsDomain0}
  on more alternative conceptions of atoms. As an example of the latter, consider autocomplete for news searches.
  News queries in Bloomberg are based on (a) a closed ontology of topics (such as oil, brexit, inflation, elections, etc.),
  tickers (unique identifiers of companies), persons, and wires (news sources); and (b) arbitrary keywords (free text).
  News queries may also specify time periods of interest, in natural language. For instance, 
  a query like {\em news about oil prices from the Financial Times last month} must be understood (by the news semantic parser) \label{Pg:NewsDomain}
  as the conjunction of \mtt{TOPIC:OIL}, \mtt{KEYWORDS:\mbox{``prices''}}, \mtt{WIRE:FT}, and \mtt{time=TimePeriod(m1/d1/y1 -- m2/d2/y2)},
  where the time period is whatever corresponds to last month. A good source of completions for news queries are news headlines,
  and more specifically, noun phrases that occur in such headlines. These noun phrases, which can be extracted with any
  standard NLP tool,
  can be viewed as a sort of atom. If one views the headlines as a query log and the noun phrases as atoms, 
  then one can extract an atom model $M$ as discussed above, with entries like:
\begin{verbatim}
"china tariffs" :=  {
  semantics := "TOPIC:CHINA & TOPIC:TARIFF",
  count := 235,
  context := {"trump": "83",
              "u.s.": "72",
              "wto": "9",
              ...
             },
  ...
 }
\end{verbatim} 
Then, given a partial query like {\em trump c}, the decomposition analysis using the news semantic parser
would understand $i_p = \temv{trump}$ as a person entity in our ontology, leaving $q_r = \temv{c}$ as an
unrecognized segment. Matching $r$ against the atom trie would pick up {\em china tariffs\/} as
a candidate completion, and then the similarity metric we outlined above would reward {\em trump china tariffs\/}
with a high score. Our news autocomplete system includes an instance of the atomic algorithm based on this approach.


\subsection{Template Completions (\mbf{template})}
\label{Sec:fqt}
The \mbf{mpc} and atomic completion schemes above both rely on access to query collections, either from users or artificially synthesized. Even in a best-case
scenario where the QA system has been deployed for a reasonably long time resulting in a large query log, it
is practically impossible to observe the entire vocabulary of a given domain in the log. 
This holds for both the semantic vocabulary (e.g. \texttt{COMPANY\_IBM}) and the natural language one (e.g. \textit{``IBM''}, 
\textit{``international business machines``}, \textit{``big blue''}). The requirement that an AC system should
be complete (Section \ref{Sec:ProblemStatement}) suggests that we also need a way to generate completions based on what the QA system can understand,
regardless of whether it has been asked before. This is where template-based completions come into play. 

Templates are an interpretable and controllable way of performing natural language generation~\cite{DBLP:conf/emnlp/WisemanSR18}. 
A template provides a schema for a natural language utterance, which can be instantiated at completion time using suitable
lexicons. Templates for natural language generation have been used before in the context of verbalizing database queries \cite{DBLP:conf/icde/KoutrikaSI10, DBLP:conf/www/NgomoBULG13}. Our setting is different in that we are not only verbalizing a logical form, but we also have
to generate the underlying semantics first with only a prefix to work with. 
Figure \ref{fqt-template-example} shows an example of a template for completion in the equities domain.

Generating fluent, well-formed natural language is an open research problem that gets harder as
utterances become longer and more compositional due to issues like agreement (on number, gender, tense) and coherence.
We have to work in this difficult setting, as our QA systems
support complex and highly compositional questions, which means that we need our AC systems to support the same. 
In fact, we want to use AC to encourage users to formulate such questions in order to utilize the full power of the Bloomberg Terminal.
There are two ways in which we can use templates for completion. The first is to use \emph{atomic templates} that capture
atoms, which are matched in a fashion similar to atomic completions (Section \ref{Sec:atomic}), but against a template rather
than a trie of atoms. The second approach is to use \emph{full-query templates} where a template specifies how complete multi-atom
queries are formulated. Multi-atom completions generated from atomic templates would need to rely on some form of scoring to handle fluency issues 
such as well-formedness, agreement and coherence. However, in contrast to the atomic completion algorithm of Section \ref{Sec:atomic}, 
which is based on data gleaned from logs, most template instances have very likely never been observed in any logs.
Additionally, even if we could somehow perform the scoring, we would need to score an impractically large number of candidates that are only materialized
at query time, which can have an unacceptable performance overhead. For these reasons we rely on full-query templates for template-based
completions. While crafting these is somewhat more time-consuming than atomic templates, we have found that the speed and completion
quality justifies the additional complexity.

In principle, a template can be instantiated offline and the resulting queries can be added to the query log 
to be used by the \mbf{mpc} and atomic completion algorithms outlined above. However, the desire to complete
compositional multi-atom queries means that we would run into a combinatorial explosion if we consider
a domain's semantic vocabulary and the corresponding lexicalizations. Because of this, we resort to instantiating
templates online at completion time in response to a user's input. Doing so gives us the additional benefit of 
being able to meaningfully deal with completions for infinite sets like numbers and dates as we detail below.

Figure \ref{fqt-template-example} shows a fragment of a full-query template template used for completing 
queries related to 
equities. Templates are encoded using the same formalism we use to encode the 
grammar used for semantic parsing in our QA systems, a cross between an algebraic
formalization of recursive augmented transition networks (ATNs)~\cite{DBLP:journals/cacm/Woods70}
and parser combinators \cite{DBLP:journals/jfp/Hutton92}, which allows us to reuse 
a great deal of the infrastructure already in place for query understanding.
Language generation in templates is primarily grounded in lexicons $\mathcal{L}$. The atomic template
$\langle\textit{enum-present}\rangle$, for example, utilizes two lexicon lookups. One is for verb phrase
aliases for fields in the present tense (such as ``trade in'') and the other is for noun aliases of values 
for these fields (e.g. ``nyse'', ``china'', but not ``chinese''). Lexicons are stored as tries for fast prefix
matching. The two lookups are connected by $\mtt{\small compatible-value-constraint()}$, a special construct 
that enforces type compatibility between the fields 
resulting from a lookup against the first trie and values from the second (`trade in', for example, 
can be paired with the exchange `nyse' but not the rating `A+'). To speed trie lookups for a subset of
values that are type-compatible with the field, $\mtt{compatible-value-constraint()}$ automatically
constructs sub-lexicons 
of $\mathcal{L}(\text{entity-value-noun})$, 
one per semantic
type, at initialization time (i.e. for locations, exchanges, ratings, etc.). Atomic templates are terminated
with the \mtt{mark} construct, used to mark the boundaries of atoms. This allows the completion algorithm
to complete atom-by-atom to the next full atom as shown in Figure~\ref{Fig:SampleNewsQueries}. 

\textbf{Lexicon Derivation.} The two lexicons used for AC in the $\langle\textit{enum-present}\rangle$ atomic template are derived
from larger lexicons used by the corresponding QA system: $\mathcal{L}(\text{entity-field})$ and 
$\mathcal{L}(\text{entity-value})$, respectively. Such derived lexicons are typically logical views of the original
ones and are not materialized, but are instead derived dynamically when the AC system is launched. 
This ensures that the QA and AC data remain in sync.
It also minimizes the amount of data that has to be maintained, allowing updates to the QA 
system to be directly reflected in the corresponding AC system.

\begin{figure*}[h]
\small

\setlength{\grammarparsep}{0.2cm}

\grammarindent1in

\noindent
\xrfill[0.7ex]{1pt}\hspace*{5mm}Primitive Templates\hspace*{5mm}\xrfill[0.7ex]{1pt}
\begin{grammar}
<logical-connectives> ::= \lit{and} | \lit{,} | \lit{or} 

<firms> ::= \lit{firms} | \lit{companies} | \lit{equities}

<numeric-pattern> ::= ( $\epsilon$ | \lit{=} | \lit{>} | \ldots | \lit{greater than} | \ldots) $\circ$ \texttt{\color{red}completable(parser = numeric-parser, sub = $\textquotestraightdblbase...\textquotestraightdblbase$)}

\end{grammar}

\noindent
\xrfill[0.7ex]{1pt}\hspace*{5mm}Atomic Templates\hspace*{5mm}\xrfill[0.7ex]{1pt}
\begin{grammar}
<enum-present> ::= $\mathcal{L}(\text{entity-field-verb-present}) \circ \texttt{\color{red}compatible-value()} \circ \mathcal{L}(\text{entity-value-noun}) \circ \texttt{\color{red}mark}$ \hspace*{3.1cm}  \textit{\color{blue}trade in nyse}

<numeric-atom> ::= $\mathcal{L}(\text{numeric-field})  \circ <numeric-pattern> \circ   \texttt{\color{red}compatible-unit()} \circ \mathcal{L}(\text{unit}) \circ \texttt{\color{red}mark}$
 \hspace*{2.2cm}  \textit{\color{blue}market cap $>$ 2... usd}
 
\end{grammar}

\noindent
\xrfill[0.7ex]{1pt}\hspace*{5mm}Full-query Templates\hspace*{5mm}\xrfill[0.7ex]{1pt}
\begin{grammar}

<adjectives> ::= $\texttt{\color{red}kleene-star-with-separator}(\mathcal{L}(\text{adjective}), <logical-connectives>)$ \hspace*{4.9cm}  \textit{\color{blue}german tech}

<display-fields> ::= $\texttt{\color{red}kleene-plus-with-separator}(\mathcal{L}(\text{display-field}), <logical-connectives>)$ \hspace*{1.7cm}  \textit{\color{blue}ipo date, ipo price and fitch rating}

<selection-query> ::= (<adjectives> $\circ$ <firms> $\circ$ \lit{with}  $\circ$ \texttt{\color{red}kleene-plus-with-separator}(<numeric-atom>, <logical-connectives>)) | ...  \\ \hspace*{8.4cm}  \textit{\color{blue}german tech companies with market cap $>$ 2... usd}

<projection-query> ::= ($<display-fields> \circ \lit{of} \circ  <firms>  \circ \lit{that} \circ \texttt{\color{red}kleene-plus-with-separator}(<enum-present>, <logical-connectives>)$) | ... \\ \hspace*{6.7cm}  \textit{\color{blue}ipo date, ipo price and fitch rating of equities that trade in nyse}

<query> ::= <selection-query> | <projection-query> | \ldots

\end{grammar}
\caption{Example of a full query template. The root of the template is \textit{query}.}
\label{fqt-template-example}
\end{figure*}

\textbf{Completing quantities with templates.} The use of templates that are instantiated online and
have full access to the expressiveness of our semantic parsing formalism results in a flexible AC framework.
We will demonstrate this by describing how we use templates to complete infinite sets like numbers. 
The problem we are tackling here is the following: As the user is typing, it is easy to complete
fields (e.g., `listed on', `market cap', `maturity date') and entities (e.g., `nyse', `bill gates', `siemens'),
as these come from finite sets, but how do we go about completing numbers,
of which there are infinitely many? One solution can be to 
hardcode a few templates with some numbers, for example:\\

\begin{small}
\noindent\hspace*{1mm}$\langle \textit{numbers} \rangle \text{::=} \text{`1000000'}| \text{`1M'} | \text{`1,000,000'}$\\
\hspace*{0mm} $\langle \textit{relation} \rangle \text{::=} \text{`='}| \text{`$>$'} | \text{`$<$'} | \ldots$\\
\hspace*{0mm} $\langle \textit{num-atom-simple} \rangle \text{::=} \mathcal{L}(\text{numeric-field}) \circ \langle \textit{relation} \rangle \circ$\\
\hspace*{30mm}$\langle \textit{numbers} \rangle \circ \mathcal{L}(\text{unit})$\\
\end{small}

\noindent The above template can complete a prefix like \textit{``market cap $>$ 1''} to \textit{``market cap $>$ 1,000,000 usd''}, but would fail
on the prefix \textit{``market cap $>$ 2''}. This failure violates the completeness requirement.  The lack of completions might (incorrectly, but understandably)
give the user the impression that their query cannot be extended to something the system can understand. This shortcoming when dealing with numbers is particularly 
unacceptable in the financial domain. While it is generally impossible to anticipate the exact number the user will type, we are mostly interested in exploiting AC to communicate
to the user that their input is a prefix of a query that we can potentially understand. A completion of the form \textit{``market cap $>$ 2... usd''} for the input \textit{``market cap $>$ 2''}, 
where the number is completed by ellipses, would communicate to the user that (i) our system understands that they are typing a number, and (ii) that the value being typed is for the \textit{market cap} field, where
\textit{usd} is an appropriate unit.

We address this problem using the \mtt{completable} construct (see the Completability algorithm in Section \ref{Sec:grammar-ac} below). 
This construct does the following: It it is initialized with a \mtt{parser} $\mathcal{P}$ (the \mtt{numeric-parser} in the $\langle \textit{numeric-pattern} \rangle$ template of Figure \ref{fqt-template-example}) and \mtt{substitution} string $sub$
(the ellipses ``\ldots'' in the same example).
At completion time, when this construct is passed a string $s$, it  checks which one of the following cases holds: 
\begin{enumerate}[i.]
  \itemsep0em 
  \item a prefix of $s$ can be  parsed by $\mathcal{P}$,
  \item $s$ is a prefix of a string that can be parsed by $\mathcal{P}$, or
  \item none of the above, indicating that no prefix of $s$ can be understood by the provided \mtt{parser}, and therefore this template cannot complete the 
       given user input.
\end{enumerate}

The following examples demonstrate what happens in each 
of the respective three cases for three different prefixes:
\begin{enumerate}[i.]
\itemsep0em 
  \item \textit{``market cap $>$ 2''} $\rightarrow$ \textit{``market cap $>$ 2... usd''}
  \item \textit{``market cap $>$ 2M u''} $\rightarrow$ \textit{``market cap $>$ 2M usd''}
  \item \textit{``market cap $>$ ibm's market c''} $\rightarrow$ template failure.
\end{enumerate}
Matching starts against the atomic template $\langle \textit{numeric-atom} \rangle$.  In all three cases ``market cap''
is matched by $\mathcal{L}(\text{numeric-field})$, with the remaining text processed by the $\langle \textit{numeric-pattern} \rangle$ template.
Here, ``$>$'' is a literal match against the first part of that template, and control moves to the \mtt{completable} construct.
We are particularly interested in the first case (i). Here $s=$``2'' is passed to the \mtt{completable} construct, and ``2'' is a prefix of a string 
that can be parsed by the \mtt{numeric-parser}. In this case, the \mtt{completable} construct produces a completion with ellipses
(\ldots) to convey to the user that the system is aware that they are currently typing a number.
The completion goes past the number to a currency unit (\textit{``usd''}) that is compatible with (\textit{``market cap''}) 
in the prefix, indicating to the user it understands that the number being typed is for a market capitalization.
The unit comes from $\mathcal{L}(\text{unit})$ back in $\langle \textit{numeric-atom} \rangle$. In (ii), $s=$\textit{``2M u''},
with the \mtt{completable} construct returning that the prefix \textit{``2M''} can be parsed as a number, and the remaining suffix \textit{``u''} is passed to the
subsequent template fragment, $\mathcal{L}(\text{unit})$. In (iii), the construct returns failure by the \mtt{numeric-parser} to  cope with any prefix of $s=$\textit{``ibm's market c''}.

\subsection{Completability Analysis}
\label{Sec:grammar-ac}

For various reasons, the above algorithms might fail to produce completions for a user's input. 
This is often by design. As explained in Section~\ref{Sec:ProblemStatement}, our QA systems 
are capable of understanding very elliptically phrased questions and commands that
are, strictly speaking, ungrammatical. We need to understand such language because it is common, 
as users have become conditioned to interacting with Web search engines and other query
interfaces using free-style keyword queries. However, producing such language in completions
is typically undesirable, as (i) the interpretation of such phrases may not be obvious to
a user reading the completion, especially to new users; and (ii) we want to use the AC
system to steer our users towards more expressive and well-structured questions. 

When none of the previously introduced algorithms return completions,
we still want to use the AC system to inform the user whether their input
is extensible to a query that the system can understand. We achieve this by a special
algorithm that essentially checks whether the underlying grammar used for semantic parsing can be used to parse
the given prefix and go just a bit beyond it, not necessarily all the way to an atom 
but simply to a complete phrase (according to the domain's vocabulary). 
No extension of the user's prefix is performed here, it is simply
a \emph{completability analysis}. The result is communicated back to the user
through the UI, informing them whether it pays off to continue with the provided prefix.
If the check fails, a user will typically go back a few characters to a point where
the system had indicated that the input is completable, and reformulate their query from there. 
A completability check will typically fail because the user is about to refer to fields,
entities, or functionality not available in a given domain. In Section \ref{Sec:fqt} we
showed how completability analysis is used to meaningfully complete inputs involving
infinite sets such as numbers.


\section{Experimental Results}
\label{Sec:ExpResults}

In this section we report on quantitative experiments intended to evaluate  
our approach to auto-completion in domains with different characteristics. 
Our experiments focus on predictiveness and efficiency. Some of the other desiderata mentioned
in Section~\ref{Sec:ProblemStatement} are guaranteed by the manner in which we compute
completions: soundness, diversity,  and propositionality. Others, like grammaticality and 
completeness (which needs to be restricted by grammaticality), are better evaluated 
qualitatively through user studies. We omit such qualitative results.

\subsection{Experimental Setup}
We present experiments in the bonds (BNDS) and news (NEWS) domains.
The two domains are markedly different, which motivates the use of 
different completion algorithms in each, and allows us to observe different
patterns in the results. In the BNDS domain, queries are issued against 
a large but  relatively static database (terminologically). The information needs are typically 
complex, as reflected by long multi-atom queries, such as:
\begin{itemize}
  \item {\em European non-gbp high yield bonds maturing between 2020 and 2030}; and
  \item {\em USD hybrid subordinate bonds issued in the last 6 months}. 
\end{itemize}
By contrast, NEWS queries are issued against news documents with various
annotations, such as publisher, publication date, 
named entities within an article (such as companies or persons), \etcsp 
A NEWS query is mapped
by our semantic parser to a combination of structured conditions matched 
against these annotations and keyword conditions matched against the body of
a document (see p.~\pageref{Pg:NewsDomain}). NEWS queries are typically
shorter than BNDS queries. Examples of NEWS queries include:
\begin{itemize}
  \item {\em ``Negative news about Guaido''} $\rightarrow$ \\\mtt{CONTAINS:PERSON_JUAN_GUAIDO AND} \\  \mtt{SEMTIMENT:NEGATIVE}
   \item {\em ``World's longest flight news from nyt''} $\rightarrow$ \\\mtt{KEYWORDS:``world's longest flight'' AND} \\  \mtt{SOURCE:NY_TIMES}
\end{itemize}
The distinguishing feature of NEWS is its dynamism. New people, topics, and keyword
phrases emerge  in the news every day, and the interests of users change accordingly.

The above characteristics motivate the algorithms used in each domain.
Because of the relatively stable state of BNDS data, the \mbf{mpc} and \mbf{atomic}
algorithms trained on reasonably sized query sets are generally sufficient.
To accommodate the dynamism of NEWS, the AC system needs to capture any semantic
entities and topics that the QA system recognizes. It also needs to be aware of the latest 
unstructured topics (e.g.,  {\em ``world's longest flight''}) 
that emerge in the news, resulting in sudden interest from users.
The first requirement is tackled by using the \mbf{template} algorithm,
which allows the AC system to stay in sync with any entities known to the
QA system, as well as the \mbf{atomic-log} algorithm, which generates
atomic-chunk completions extracted offline from queries. 
The second requirement, related to unstructured topics, is tackled
by using an instance of the atomic algorithm, \mbf{atomic-headline}, 
that is trained on noun phrases extracted from news documents (p.~\pageref{Pg:NewsDomain0}).

Our experimental setup is intended to test how robustly predictive
our AC systems are. To do so, we simulate users interacting with 
our systems to ask {\em questions that have never been asked before}.
We do so by taking an existing set of queries $Q$ from a time interval
$[t_1, t_2]$, specifying a  a cutoff time $t_{\text{cutoff}}$ in that interval,
and partitioning $Q$ into two sets $Q_1$ and $Q_2$, respectively comprising the queries
that appear in $[t_1, t_{\text{cutoff}}]$ and  $(t_{\text{cutoff}}, t_2]$. 
$Q_1$ will be used to train those algorithms that need training as described below.
Now, to simulate users asking questions that the system has not seen before, 
we use $Q_2 \setminus Q_1$ as our set of test queries (where $\setminus$ is the set difference operator).
Because training and testing queries are disjoint, \mbf{mpc} is effectively 
useless and is taken out of consideration for both BNDS and NEWS. 
For BNDS, this leaves the \mbf{atomic} algorithm trained on $Q_1$.
For NEWS, $Q_1$ is used to train \mbf{atomic-log}. Additionally,
\mbf{atomic-headline} is trained on news headlines from articles
published up until $t_2$, and \mbf{template} uses the data available to the corresponding NEWS QA system.

Following previous 
work, we restrict ourselves to prefixes of length at least 3 characters, when a word starts to emerge,
as the completion of very short prefixes like `c' and `ci' is somewhat ill-defined.
The AC systems are configured to return 10 completions for a given prefix.
Individual algorithms will typically generate many more 
completion candidates, allowing the top-level coordinating algorithm (Section \ref{Sec:Approaches})
to choose the best 10 (accounting for grades, the need for diversification, and so on). 
Completion algorithms run concurrently, and pass their results to the top-level
algorithm to be merged.

\subsection{Predictiveness}\label{predictiveness-exp}
A predictive AC system is one that can suggest a user's intended query 
closer to the top of the completion list. In this section
we present experiments that demonstrate the predictive abilities of our AC 
approach. 

\textbf{Predictiveness Measures}. 
We evaluate predictiveness using the standard approach in the literature:
For each test query $q = [c_1,...,c_n]$ (where each $c_i$ is a character), we generate its  
prefixes, each one of the form $p^j_q = [c_{1},...,c_{j}]$, $j \in \{1,\ldots,n\}$.
For each prefix, we would like  a completion that matches $q$, in some appropriate sense,
to appear as high as possible in the completion list produced by the AC system.
Let $\temv{completions}(p^j_q) = [q'_1,...,q'_m]$ be the 
ordered list of completions returned by the AC system, and let 
 $\temv{match}(q, q') \in \{0,1\}$
be a predicate that returns 1 if $q$ and $q'$ have the same intent and
0 otherwise. We will present several possible instantiations of this predicate below.
The reciprocal rank (RR) is defined as follows:\[RR(q,\temv{completions}(p^j_q)) = \frac{1}{min(\{i \sep match(q, q'_i) = 1\})}\] 
For an evaluation query collection $Q$, we report the Mean Reciprocal Rank (MRR), which is the 
average of all RR scores over all prefixes of all queries.


\begin{table}[t!]
  \renewcommand{\arraystretch}{1.2}
  \centering
  \caption{Predictiveness}
  
  \begin{tabular}{| l | c | c || c | c |} 
      \hline
        & \multicolumn{2}{|c||}{BNDS} & \multicolumn{2}{|c|}{NEWS}  \\ 
      \hline
      $\temv{match}(q, q')$ & \multirow{2}{*}{MRR}  & {PARTIAL}  & \multirow{2}{*}{MRR}  &{PARTIAL} \\ 
      instantiation & & MRR & & MRR \\
      \hline
      STR            & 0.028 &  0.374 &     0.226           &     0.355          \\
      BOW            & 0.031 &  0.442 &    0.243           &     0.457          \\
      SEM            & 0.081 &  0.589 &    0.256           &     0.491           \\  
      \hline
  \end{tabular}
  \label{predictiveness-result}
\end{table}

We now present various instantiations of the $\temv{match}(q, q')$ predicate above.
The simplest one, STR, is when we look for exact string matches, i.e., $q = q'$. 
A more appropriate measure for settings like ours, where queries can be long, 
has been introduced by Park and Chiba~\cite{DBLP:conf/sigir/ParkC17}, and is known as
the {\em partial\/} match criterion (PSTR). Here the completion can be the same as 
$q$ or a prefix of it. Partial matching is an important notion in our setting due to the 
propositionality of our completions (Section \ref{Sec:ProblemStatement}). Outside of \mbf{mpc},
our core AC algorithms are designed to complete to the next atom, so for a prefix like $p_q^{18}${\em european non-gbp h}
originating from the reference query $q =$ {\em european non-gbp high yield bonds maturing between 2020 and 2030}, 
our systems would generate $q' =$ {\em european non-gbp high yield bonds}, which is a match under PSTR
but not under STR.

Next, if we look at the definition of syntactic soundness  in Section \ref{Sec:ProblemStatement},
a valid completion can reorder the individual tokens in the user's input prefix. For example, a valid completion for ``guai'' is 
``juan guaido''. This easily extends to longer multi-atom questions. To capture this, we generalize the STR matching to 
bag-of-word (BOW) matching, where a match occurs if the set of words in $q$ and the completion are the same. This can 
also be relaxed in the same way that PSTR is a relaxation of STR, through the notion of the bag-of-word subset (PBOW).

Finally, given that we are in a setting where we have access to both completions and their semantic interpretations, 
we can perform matching based on the semantics rather than the surface form of the completion, resulting in the 
SEM and PSEM matching predicates. For example, consider an input query \temv{ibm} in the BNDS domain. The
semantics of this query are captured by a formula along the lines of
\begin{equation}
  \mtt{ISSUING\_COMPANY} = \mtt{COMPANY\_IBM}.
  \label{Eq:CompSem}
  \end{equation}
Given a prefix like \temv{ib}, the completion \temv{ibm bonds} is, strictly speaking, neither a prefix of
the intended query (\temv{ibm}), nor the same bag of words.  Accordingly, neither of the partial measures described
above would count this completion as successful. However, the semantics of the completion are in fact identical
to the semantics of the intended query, as given by~(\ref{Eq:CompSem}), and by that important measure,
the completion is in fact a perfect match, even if its lexical form is different. 

Table \ref{predictiveness-result} shows the results of the predictiveness experiment.
It is important to keep in mind that in our setting we are completing to queries that have never been 
observed before (by restricting evaluation to $Q_2 \setminus Q_1$). 
With this in mind, we start by analyzing the results for BNDS. The contrast between (full match) MRR
and partial match MRR is striking. The low MRR numbers are expected given the highly compositional 
nature of BNDS queries and the fact that the algorithm we are testing is designed to complete to the next
full atom, which makes a complete match only possible when the prefix contains characters for the very
last atom in the reference query. If we focus on the partial MRR results for BNDS, we can observe how,
as we move down the table, the numbers increase, reflecting the fact that our AC systems are semantically
driven, which means that even if they do not reproduce the exact same reference query, they produce
a semantic equivalent with possible reordering of words. The semantic (SEM) partial MRR reflects
that the desired completion appears, on average, at rank 1 or 2.

The MRR numbers for NEWS reflect the contrast between query length and the degree of query
compositionality compared to BNDS. As in BNDS, completion in NEWS is done to the next atom. However,
since the queries are shorter, exact matches are easier to come across. The partial MRR numbers for news
tell the complete picture about the quality of the results. They are in line with the numbers observed for
BNDS, with the desired completion appearing at rank 2 on average, which is a strong result
given that none of the target queries has been seen in its entirety by the system before.

\subsection{Efficiency}

AC systems need to be highly responsive: We impose a hard upper bound of 100ms between
a user's keystroke and the presentation of the corresponding completions on the screen,
in order to ensure interactivity and avoid the user noticing any lag \cite{DBLP:conf/afips/Miller68}. 
This time span needs to include not only the time it takes to compute completions,
but also the time needed to transmit them over a network and paint them on 
the UI. We have been targeting response times of well below 100ms, and 
our experiments demonstrate that we consistently achieve them. 

We report on the mean completion time, as well as the $n$th percentile
$P_{n}$ for several values of $n$. Table \ref{ac-timing-results} shows the results we obtain.
All numbers indicate that our AC systems are fast, even when considering 
the 99th percentile, we are well below the 100ms maximum allotted for the end-to-end
completion of a prefix. 

\begin{table}[h]
 \renewcommand{\arraystretch}{1.3}
  \centering
  \caption{Response times in ms.}
  \begin{tabular}{| l | c | c | c | c | c | c |} 
      \hline
        System & Mean & $P_{90}$ & $P_{95}$ & $P_{99}$  \\
       \hline
        NEWS &  11.37   &   22.21       &   27.46       &    42.51              \\      
        BNDS & 6.17   &   9.31       &   14.35      &    45.93              \\       

      \hline
  \end{tabular}
\label{ac-timing-results}
\end{table}

The \mbf{template} algorithm is generally the most time-consuming algorithm, the reason being that, 
compared to other algorithms, \mbf{template} generates its completions exclusively online, which
means that it cannot cache metadata for these completions like other algorithms do.
This translates
into \mbf{template} needing to semantically analyze the candidates it produces (typically between 20 and 100)
on the fly, at completion time. The \mbf{atomic} algorithm needs to perform one parsing operation of the user's input prefix to determine the boundary of the
atom it should complete. All other operations are based on metadata that is generated offline and cached for speed. 
\mbf{mpc}, which is not part of our experiments here, is an order of magnitude faster than the above two algorithms (sub-millisecond) as 
it fully relies on pre-computed metadata. As demonstrated by the above numbers, all algorithms meet the
responsiveness constraints.


\section{Related Work}
\label{Sec:RelatedWorkAndConclusions}
We developed the AC framework described in this work and the corresponding
QA framework in the context of the larger problem of \emph{improving the usability of information systems}.
These usability issues have long been recognized 
\cite{DBLP:conf/sigmod/JagadishCEJLNY07,DBLP:journals/debu/LiJ12}. Our focus here is
on usable query interfaces. A wide array of solutions have been proposed, from visual query interfaces \cite{DBLP:journals/pvldb/BhowmickCD16}
to textual ones based on keyword queries  \cite{DBLP:conf/vldb/AdityaBCHNS02, DBLP:conf/emnlp/JoshiSC14, DBLP:conf/vldb/YuJ07} to 
interfaces based fully on natural language\cite{DBLP:conf/sigmod/LiJ14, DBLP:journals/sigmod/LiJ16, DBLP:conf/vldb/YuJ07}, 
like the ones we have been developing. Semantic parsing for question answering
has a long history \cite{18102330,DBLP:journals/cacm/Woods70}. It has seen a revival in recent years \cite{DBLP:journals/pvldb/LiJ14}
brought on in part by business needs whereby non-technical users need access to expressive query interfaces, 
as well as the proliferation of smart phones and digital personal assistants, where natural language is a convenient mode 
of interaction. 

We tackle the AC problem from a unique angle, heavily informed by the semantics of the underlying
query and with the aim of supporting semantic query interfaces. Auto-completion, however, has a long history in the
information retrieval community for search interfaces over text documents \cite{INR055}.
Most work in that setting is 
based on completion from query logs using various flavors of \mbf{mpc}, with the main focus being on appropriate 
ranking of completions~\cite{DBLP:conf/sigir/Shokouhi13}. 
An important aspect in ranking completions is 
time-sensitivity~\cite{DBLP:conf/cikm/CaiLR14}, reflecting dynamic user needs. This is an aspect that our
algorithms have also given attention to in time-sensitive domains like news.
Work has also been done on providing completions in the absence of query logs in relatively small-scale search
settings like enterprise, intranet, and email search~\cite{DBLP:conf/sigir/BastW06}.
In this setting, completions are generated either by means of phrase extraction and
scoring from the underlying corpora~\cite{DBLP:conf/sigir/BhatiaMM11,DBLP:conf/sigir/HorovitzLLMR17}. In Section
\ref{Sec:atomic} we described how we use a similar approach in the \mbf{atomic} algorithm
for the News domain by relying on phrases extracted from news article headlines.
More recent work has looked at completing before-unseen prefixes by generating synthetic
completions based on $n$-grams extracted from query logs, and relying on neural ranking methods
due to their ability to generalize~\cite{DBLP:conf/cikm/MitraC15}. This is somewhat
similar to our \mbf{atomic}  algorithm,  the main distinction being that instead of arbitrary $n$-grams
we rely on semantics to complete to full atoms.

Auto-completion has also been explored for formal query languages like SQL \cite{DBLP:journals/pvldb/KhoussainovaKBS11} and SPARQL 
\cite{DBLP:conf/cikm/BastB17}. The goal here is to help users formulate correct formal queries in cases where they might be
unfamiliar with language constructs or the vocabulary of the underlying data. Like ours,  these
systems are guided by semantics and strive to make contextually relevant suggestions.  AC for formal query languages
can be very helpful for technically proficient users. Our QA and corresponding AC systems
target a wider user base, where such proficiency cannot be assumed.

Another line of related work is the verbalization of formal queries \cite{DBLP:conf/icde/KoutrikaSI10, DBLP:conf/www/NgomoBULG13}.
The primary goal here is  to allow users to confirm that the formal queries they typed, or those produced by 
a form-filling interface, capture their information needs.
Verbalization systems typically rely on templates, which we also use. These
systems take a complete formal query and produce a verbalization. 
The problem we tackle is a somewhat more challenging one where we're given only 
a natural language prefix and we effectively have to predict the semantic intent (the formal query) and
then verbalize it in a manner consistent with the input prefix.




\begin{thebibliography}{10}

\bibitem{DBLP:conf/vldb/AdityaBCHNS02}
B.~Aditya, G.~Bhalotia, S.~Chakrabarti, A.~Hulgeri, C.~Nakhe, Parag, and
  S.~Sudarshan.
\newblock {BANKS: Browsing and Keyword Searching in Relational Databases}.
\newblock In {\em {VLDB}}, pages 1083--1086, 2002.

\bibitem{DBLP:conf/cikm/BastB17}
H.~Bast and B.~Buchhold.
\newblock {QLever: A Query Engine for Efficient SPARQL+Text Search}.
\newblock In {\em {CIKM}}, pages 647--656, 2017.

\bibitem{DBLP:conf/cikm/BastH15}
H.~Bast and E.~Haussmann.
\newblock {More Accurate Question Answering on Freebase}.
\newblock In {\em {CIKM} 2015}, pages 299--304, 2015.

\bibitem{DBLP:conf/sigir/BastW06}
H.~Bast and I.~Weber.
\newblock Type less, find more: fast autocompletion search with a succinct
  index.
\newblock In {\em {SIGIR}}, pages 364--371, 2006.

\bibitem{DBLP:conf/sigir/BhatiaMM11}
S.~Bhatia, D.~Majumdar, and P.~Mitra.
\newblock Query suggestions in the absence of query logs.
\newblock In {\em {SIGIR}}, pages 795--804, 2011.

\bibitem{DBLP:journals/pvldb/BhowmickCD16}
S.~S. Bhowmick, B.~Choi, and C.~E. Dyreson.
\newblock {Data-driven Visual Graph Query Interface Construction and
  Maintenance: Challenges and Opportunities}.
\newblock {\em {PVLDB}}, 9(12):984--992, 2016.

\bibitem{INR055}
F.~Cai and M.~de~Rijke.
\newblock {A Survey of Query Auto Completion in Information Retrieval}.
\newblock {\em Foundations and Trends in Information Retrieval},
  10(4):273--363, 2016.

\bibitem{DBLP:conf/cikm/CaiLR14}
F.~Cai, S.~Liang, and M.~de~Rijke.
\newblock {Time-sensitive Personalized Query Auto-Completion}.
\newblock In {\em {CIKM}}, pages 1599--1608, 2014.

\bibitem{Helander:1997:HHI:549940}
M.~G. Helander, T.~K. Landauer, and P.~V. Prabhu, editors.
\newblock {\em Handbook of Human-Computer Interaction}.
\newblock Elsevier Science Inc., New York, NY, USA, 2nd edition, 1997.

\bibitem{DBLP:conf/sigir/HorovitzLLMR17}
M.~Horovitz, L.~Lewin{-}Eytan, A.~Libov, Y.~Maarek, and A.~Raviv.
\newblock Mailbox-based vs. log-based query completion for mail search.
\newblock In {\em SIGIR}, 2017.

\bibitem{DBLP:journals/jfp/Hutton92}
G.~Hutton.
\newblock {Higher-Order Functions for Parsing}.
\newblock {\em Journal of Functional Programming}, 2(3):323--343, 1992.

\bibitem{DBLP:conf/sigmod/JagadishCEJLNY07}
H.~V. Jagadish, A.~Chapman, A.~Elkiss, M.~Jayapandian, Y.~Li, A.~Nandi, and
  C.~Yu.
\newblock {Making Database Systems Usable}.
\newblock In {\em {SIGMOD}}, pages 13--24, 2007.

\bibitem{DBLP:conf/emnlp/JoshiSC14}
M.~Joshi, U.~Sawant, and S.~Chakrabarti.
\newblock {Knowledge Graph and Corpus Driven Segmentation and Answer Inference
  for Telegraphic Entity-seeking Queries}.
\newblock In {\em {EMNLP}}, pages 1104--1114, 2014.

\bibitem{DBLP:journals/corr/abs-1812-00978}
A.~Kamath and R.~Das.
\newblock A survey on semantic parsing.
\newblock {\em CoRR}, abs/1812.00978, 2018.

\bibitem{DBLP:journals/pvldb/KhoussainovaKBS11}
N.~Khoussainova, Y.~Kwon, M.~Balazinska, and D.~Suciu.
\newblock {SnipSuggest: Context-Aware Autocompletion for SQL}.
\newblock {\em {PVLDB}}, 4(1):22--33, 2010.

\bibitem{DBLP:conf/icde/KoutrikaSI10}
G.~Koutrika, A.~Simitsis, and Y.~E. Ioannidis.
\newblock {Explaining Structured Queries in Natural Language}.
\newblock In {\em {ICDE}}, pages 333--344, 2010.

\bibitem{DBLP:journals/debu/LiJ12}
F.~Li and H.~V. Jagadish.
\newblock {Usability, Databases, and HCI}.
\newblock {\em {IEEE} Data Engineering Bulletin}, 35(3):37--45, 2012.

\bibitem{DBLP:journals/pvldb/LiJ14}
F.~Li and H.~V. Jagadish.
\newblock {Constructing an Interactive Natural Language Interface for
  Relational Databases}.
\newblock {\em {PVLDB}}, 8(1):73--84, 2014.

\bibitem{DBLP:conf/sigmod/LiJ14}
F.~Li and H.~V. Jagadish.
\newblock {NaLIR: an Interactive Natural Language Interface for Querying
  Relational Databases}.
\newblock In {\em {SIGMOD}}, pages 709--712, 2014.

\bibitem{DBLP:journals/sigmod/LiJ16}
F.~Li and H.~V. Jagadish.
\newblock {Understanding Natural Language Queries over Relational Databases}.
\newblock {\em {SIGMOD} Record}, 45(1):6--13, 2016.

\bibitem{DBLP:conf/afips/Miller68}
R.~B. Miller.
\newblock {Response Time in Man-Computer Conversational Transactions}.
\newblock In {\em {AFIPS}}, pages 267--277, 1968.

\bibitem{DBLP:conf/cikm/MitraC15}
B.~Mitra and N.~Craswell.
\newblock {Query Auto-Completion for Rare Prefixes}.
\newblock In {\em {CIKM}}, pages 1755--1758, 2015.

\bibitem{DBLP:conf/www/NgomoBULG13}
A.~N. Ngomo, L.~B{\"{u}}hmann, C.~Unger, J.~Lehmann, and D.~Gerber.
\newblock {Sorry, I Don't Speak SPARQL: Translating SPARQL Queries into Natural
  Language}.
\newblock In {\em {WWW}}, pages 977--988, 2013.

\bibitem{DBLP:conf/sigir/ParkC17}
D.~H. Park and R.~Chiba.
\newblock {A Neural Language Model for Query Auto-Completion}.
\newblock In {\em {SIGIR}}, pages 1189--1192, 2017.

\bibitem{DBLP:conf/acl/SavenkovA17}
D.~Savenkov and E.~Agichtein.
\newblock {EviNets: Neural Networks for Combining Evidence Signals for Factoid
  Question Answering}.
\newblock In {\em {ACL}}, pages 299--304, 2017.

\bibitem{DBLP:conf/sigir/Shokouhi13}
M.~Shokouhi.
\newblock {Learning to Personalize Query Auto-Completion}.
\newblock In {\em {SIGIR} 2013}, pages 103--112, 2013.

\bibitem{DBLP:conf/emnlp/WisemanSR18}
S.~Wiseman, S.~M. Shieber, and A.~M. Rush.
\newblock {Learning Neural Templates for Text Generation}.
\newblock In {\em {EMNLP}}, pages 3174--3187, 2018.

\bibitem{18102330}
W.~Woods, R.~Kaplan, and B.~Nash-Webber.
\newblock {The Lunar Sciences Natural Language Information System}.
\newblock Technical report, BBN Inc., 1974.

\bibitem{DBLP:journals/cacm/Woods70}
W.~A. Woods.
\newblock {Transition Network Grammars for Natural Language Analysis}.
\newblock {\em Communications of the {ACM}}, 13(10):591--606, 1970.

\bibitem{DBLP:conf/vldb/YuJ07}
C.~Yu and H.~V. Jagadish.
\newblock {Querying Complex Structured Databases}.
\newblock In {\em {VLDB}}, pages 1010--1021, 2007.

\end{thebibliography}




\end{document}